\PassOptionsToPackage{unicode=true}{hyperref} 
\PassOptionsToPackage{hyphens}{url}
\PassOptionsToPackage{dvipsnames,svgnames*,x11names*}{xcolor}
\documentclass[12pt,]{article}
\usepackage{lmodern}
\usepackage{amssymb,amsmath}
\usepackage{ifxetex,ifluatex}
\usepackage{fixltx2e} 
\ifnum 0\ifxetex 1\fi\ifluatex 1\fi=0 
  \usepackage[T1]{fontenc}
  \usepackage[utf8]{inputenc}
  \usepackage{textcomp} 
\else 
  \usepackage{unicode-math}
  \defaultfontfeatures{Ligatures=TeX,Scale=MatchLowercase}
\fi
\IfFileExists{upquote.sty}{\usepackage{upquote}}{}
\IfFileExists{microtype.sty}{%
\usepackage[]{microtype}
\UseMicrotypeSet[protrusion]{basicmath} 
}{}
\IfFileExists{parskip.sty}{%
\usepackage{parskip}
}{
\setlength{\parindent}{0pt}
\setlength{\parskip}{6pt plus 2pt minus 1pt}
}
\usepackage{xcolor}
\usepackage{hyperref}
\hypersetup{
            colorlinks=true,
            linkcolor=Maroon,
            filecolor=Maroon,
            citecolor=Blue,
            urlcolor=blue,
            breaklinks=true}
\usepackage[margin=1.0in]{geometry}
\usepackage{graphicx,grffile}
\makeatletter
\def\maxwidth{\ifdim\Gin@nat@width>\linewidth\linewidth\else\Gin@nat@width\fi}
\def\maxheight{\ifdim\Gin@nat@height>\textheight\textheight\else\Gin@nat@height\fi}
\makeatother
\setkeys{Gin}{width=\maxwidth,height=\maxheight,keepaspectratio}
\providecommand{\tightlist}{%
  \setlength{\itemsep}{0pt}\setlength{\parskip}{0pt}}
\ifx\paragraph\undefined\else
\let\oldparagraph\paragraph
\renewcommand{\paragraph}[1]{\oldparagraph{#1}\mbox{}}
\fi
\ifx\subparagraph\undefined\else
\let\oldsubparagraph\subparagraph
\renewcommand{\subparagraph}[1]{\oldsubparagraph{#1}\mbox{}}
\fi

\makeatletter
\def\fps@figure{htbp}
\makeatother

\usepackage{longtable}
\usepackage{graphicx}
\usepackage{array}
\usepackage{caption}
\usepackage{graphicx}
\usepackage{siunitx}
\usepackage{ulem}
\usepackage{colortbl}
\usepackage{multirow}
\usepackage{hhline}
\usepackage{calc}
\usepackage{tabularx}
\usepackage{threeparttable}
\usepackage{wrapfig}
\usepackage{adjustbox}
\usepackage{hyperref}

\author{}
\date{\vspace{-2.5em}}

\begin{document}

\pagenumbering{gobble}

\begin{centering}

$ $

\vspace{5cm}

\LARGE

{\bf Interpretable similarity-driven multi-view embeddings from high-dimensional biomedical data}

\vspace{1.0 cm}

\normalsize

Brian B. Avants, PhD,
Nicholas J. Tustison, DSc,
James R. Stone, MD, PhD, and
the Pediatric Imaging, Neurocognition, and Genetics Study$^*$

Department of Radiology and Medical Imaging, University of Virginia, Charlottesville, VA

\end{centering}

\small

\vspace{3.0 cm}

Corresponding author:\\
Brian B. Avants, PhD\\
Department of Radiology and Medical Imaging\\
University of Virginia\\
480 Ray C Hunt Drive\\
Box 801339\\
Charlottesville, VA 22903\\
434-924-9585\\
\href{mailto:stnava@gmail.com}{\nolinkurl{stnava@gmail.com}}

\noindent

\rule{4cm}{0.4pt}

\footnotesize

*Data used in preparation of this article were obtained from the
Pediatric Imaging, Neurocognition and Genetics Study (PING) database
(\url{http://ping.chd.ucsd.edu}). As such, the investigators within PING
contributed to the design and implementation of PING and/or provided
data but did not participate in analysis or writing of this report. A
complete listing of PING investigators can be found at
\url{https://ping-dataportal.ucsd.edu/sharing/Authors10222012.pdf}

\normalsize

\newpage

\textbf{Abstract}

Similarity-driven multi-view linear reconstruction (SiMLR) is an
algorithm that exploits inter-modality relationships to transform large
scientific datasets into smaller, more well-powered and interpretable
low-dimensional spaces. SiMLR contributes a novel objective function for
identifying joint signal, regularization based on sparse matrices
representing prior within-modality relationships and an implementation
that permits application to joint reduction of large data matrices, each
of which may have millions of entries. We demonstrate that SiMLR
outperforms closely related methods on supervised learning problems in
simulation data, a multi-omics cancer survival prediction dataset and
multiple modality neuroimaging datasets. Taken together, this collection
of results shows that SiMLR may be applied with default parameters to
joint signal estimation from disparate modalities and may yield
practically useful results in a variety of application domains.


\clearpage

Healthcare -- from both a prevention as well as treatment perspective --
is increasingly turning to large, mixed datasets to gain a better
understanding of the biological complexity that influences sensitivity
or resistance to disease, injury, etc. In the case of rare diseases,
multi-view datasets are collected to build a more complete
characterization of disease phenotype and potentially gain insights into
etiology. In more common conditions, like Alzheimer's disease,
multi-view datasets are motivated by the need to understand the
diversity of the disease process, identify sub-groups and thereby
advance personalized treatment approaches. Multi-view data can also
provide insight into the features that drive variability within the
``normal'' phenotype e.g.~underlying factors that contribute to
difference in neurobiological age\textsuperscript{1} or the genetic
architecture of quantitative phenotypes as mediated through brain
structure\textsuperscript{2}.

Multi-view (also known as multiple modality or multi-block) datasets are
increasingly common in the biomedical sciences. In the idealized case,
each view / modality will provide a completely unique measurement of the
substrate biology. However, it is perhaps more common that each view
provides a partial and not wholly independent perspective on a complex
phenomenon. In this case, covariation can be exploited in order to sift
through noisy measurements and better identify meaningful signal.
Moreover, joint relationships across systems of the brain or across
scale can form the foundation for integrative scientific hypotheses.

Pre-specified joint hypotheses allow the scientist to avoid a
combinatorial explosion of tests for possible interactions. Although
powerful in sufficiently large, well-understood datasets, prior
multivariate hypotheses can be difficult to enumerate with sufficient
detail to support implementation and testing. Fully multivariate and
data-driven dimensionality reduction models provide an alternative
including principal component analysis (PCA)\textsuperscript{3,4} and
independent component analysis (ICA)\textsuperscript{5--7}. However,
these popular models, applied directly, are not explicitly designed for
interpretation across multiple modalities and do not provide an easy way
for the scientist to regularize the solution with prior knowledge or to
visualize the feature vectors which are both dense and signed (i.e.~have
both positive and negative weights).

Graph-regularized, imaging-focused dimensionality reduction methods
emerged in recent years to address the desire for interpretable
components\textsuperscript{8--10}.
\texttt{Graph-net}\textsuperscript{10}, similar to
\texttt{SCCAN}\textsuperscript{11,12}, uses \(\ell_1\) regularization to
constrain embedding vectors to be sparse and reduce over-fitting in
high-dimensional problems. Relatedly, graph-regularization has been used
to improve prediction in imaging genetics\textsuperscript{10,13,14} and
may be combined with canonical correlation analysis (as in
\texttt{SCCAN}\textsuperscript{11}). Non-negative factorization methods
provide a second degree of interpretability by guaranteeing that
factorizations are unsigned and, therefore, allow components to be
interpreted in terms of their original units
(e.g.~millimeters)\textsuperscript{15,16}. Other
efforts\textsuperscript{9,17} use prior constraints to guide solutions
toward familiar sparsity patterns. More generally, regularization is
also critical to well-posedness\textsuperscript{18,19}.

The need for joint, interpretable modeling of several (\textgreater{}2)
parallel but heterogenous datatypes is rapidly
increasing\textsuperscript{20--24}. Multi-block data analysis methods
such as Kettering's five offerings\textsuperscript{25} and more recent
regularized generalized canonical correlation analysis (RGCCA and its
sparse variant SGCCA)\textsuperscript{26--28} and multiway generalized
canonical correlation analysis\textsuperscript{29} extend Hotelling's
classical CCA\textsuperscript{30,31} to multi-view (viz. multi-block)
data. Joint and individual variation explained (JIVE) is another
framework dedicated to data fusion\textsuperscript{32,33} along with
MultiLevel Simultaneous Component Analysis (MLSCA)\textsuperscript{34}
and Multi‐Omics Factor Analysis (MOFA)\textsuperscript{35}. A variation
of JIVE, applied to convolutional network features, has also been
applied to imaging genetics problems\textsuperscript{36}.

Our contribution, similarity-driven multi-view linear reconstruction
(SiMLR), is a new joint embedding method---targeting biomedical
data---that links several of the ideas expressed in prior work. SiMLR
builds on sparse canonical correlation analysis for neuroimaging
(\texttt{SCCAN})\textsuperscript{12,37,38} and prior-based
eigenanatomy\textsuperscript{17,39}. SiMLR goes beyond \texttt{SCCAN} in
that it takes two or more modalities as input, allows customized
regularization models and uses a fast and memory efficient
implementation appropriate for large datasets. SiMLR outputs locally
optimal low-dimensional matrix embeddings for each modality that best
predict its partner modalities. SiMLR achieves this by reconstructing
each modality matrix from a basis set derived from the partner
modalities. One novel aspect of SiMLR is that the ``linking'' basis set
is computed via a source separation algorithm, e.g.~singular value
decomposition (SVD) or ICA. This sub-algorithm seeks to identify latent
signal sources that span modalities. The basis set can be forced to be
either orthogonal (SVD) or statistically independent (ICA) where the
latter option may be more appropriate for unmixing signal sources in
real world data\textsuperscript{40,41}. Simultaneously, the feature
vectors may be constrained by graph-regularized sparsity and
non-negativity. Furthermore, the target energy (measuring the similarity
between different modalities) is also flexible and builds on classical
objective functions in SVD and CCA. SiMLR is the only available
framework that combines these features in an accessible and flexible
joint dimensionality reduction algorithm. \footnote{Although SiMLR
supports path modeling, only the leave-one-modality-out
  approach is explored in this work.}

\hypertarget{results}{%
\section{Results}\label{results}}

Figure 1 shows a general overview of how SiMLR is applied within the
context of scientific data. Each evaluation below fits within this
general framework. Furthermore, each study uses joint dimensionality
reduction in conjunction with regression-based supervised learning in a
training and testing paradigm. Table 1 summarizes the overall findings
that are presented in this results section and demonstrates a ranking of
performance based on the included evaluation metrics. The results below
are all drawn from the following location where the underlying
reproducible computational practices are made available to readers:
\url{https://codeocean.com/capsule/9877797/}. From here, this location
will be referred to simply as \texttt{codeocean} and specific files
therein will be referenced by filename as they appear in the
\texttt{results} section of the capsule.

\hypertarget{terminology}{%
\subsection{Terminology}\label{terminology}}

\noindent We outline the terminology used in the discussion that
follows.

\begin{itemize}
\item
  \textbf{Multi-view:} several modalities collected in one cohort;
  alternatively, the same measurements taken across different
  studies\textsuperscript{42}. We focus on the first case here.
\item
  \textbf{Covariation:} we use the term in two contexts. As a general
  concept, we mean systematic changes in one modality are reflected in a
  predictable amount of change in other modalities. In the mathematical
  context, we use the definition of covariation for discrete random
  variables.
\item
  \textbf{Latent space/embeddings:} both terms refer to an (often
  lower-dimensional) representation of high-dimensional data. These are
  also known as components in PCA. In the context of this paper, we are
  \emph{approximating} the (hidden) latent space with the learned
  embeddings. Often, the true latent space cannot be known. We compute
  embeddings (or components), here, by multiplying feature vectors
  against input data matrices. Importantly, SiMLR can compute latent
  spaces that target either statistical independence (the ICA source
  separation algorithm\textsuperscript{40}) or orthogonality (the SVD
  algorithm). Deflation-based schemes, on the other hand, only target
  orthogonality.
\item
  \textbf{Feature vectors:} these are weights on the original features.
  In SiMLR, the feature vectors are the solutions that we are seeking.
  Projecting the feature vectors onto the original data will provide a
  low-dimensional representation.
\end{itemize}

These concepts are expanded upon in more detail in the methodology
section.

\hypertarget{data-representation}{%
\subsection{Data representation}\label{data-representation}}

SiMLR assumes ``clean'' data as input. This data has no missing values
and is structured in matrix format with each modality matched along rows
(here, these are the subjects/samples). Single nucleotide polymorphism
(SNP) data is often formatted this way after imputing to a common
reference dataset such as the HapMap. In neuroimaging, we employ region
of interest measurements or spatial normalization in order to map a
high-dimensional image into this common representation. For example, if
a brain template has \(p\) voxels within the cortex and the population
contains \(n\) subjects, then the matrix representation of the
population level voxel-wise, normalized cortical thickness map will be
\(X_\text{thickness}\) with dimensions \(n \times p\). SiMLR accepts
\(>1\) matrices organized in this manner. A study of \(m\) distinct
modalities would have input matrices with dimensions \(n \times p_i\)
(subjects \(\times\) predictors), noting that \(p_i\) need not equal
\(p_j\) for any \(i, j \in {1,\cdots,m}\).

\newpage

\begin{figure}
\includegraphics[width=1\linewidth]{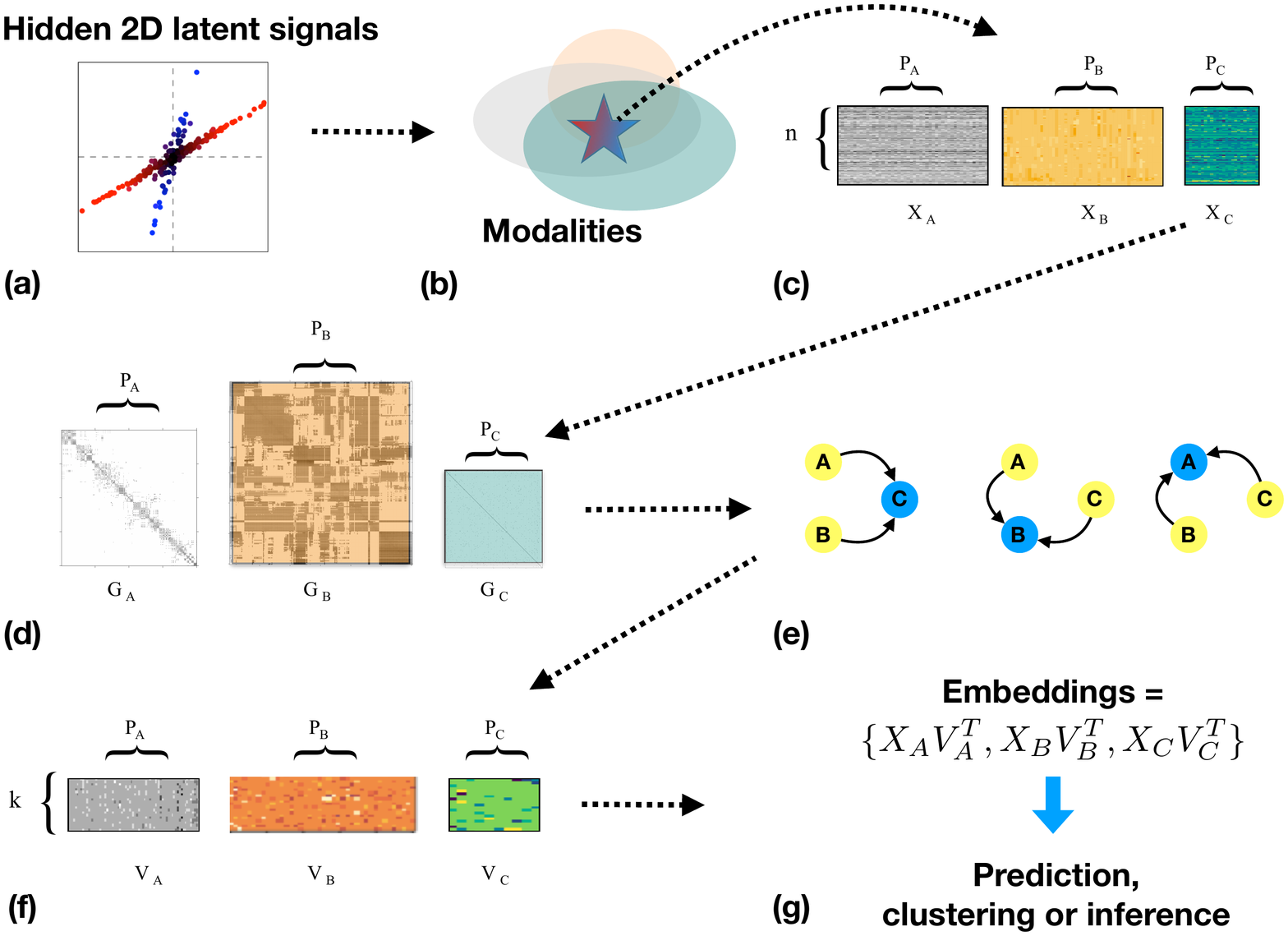} \caption{Example SiMLR study overview: (a) Two statistically independent signals are shown here to represent the hidden latent signal potentially two components of a disease process; (b) This latent signal is distributed across different subjects each of whom is measured with multi-omics. (c) The data is converted to matrices; in this effort we focus on matrices with common number of subject here denoted by $n$ and variable number of predictors ($p_{A,B,C}$). (d) Sparse regularization matrices ($G$) are constructed with user input of domain knowledge or via helper functions; (e) SiMLR iteratively optimizes the ability of the modalities to predict each other in leave one out fashion; (f) Sparse feature vectors emerge which can be interpreted as weighted averages over select columns of the input matrices that maintain the original units of the data.  These are used to compute embeddings in (g) and passed to downstream analyses.  Alternatively, one could permute the SiMLR solution to gain empirical statistics on its solutions.}\label{fig:fig1}
\end{figure}

\newpage

\begin{figure}
\includegraphics[width=1\linewidth]{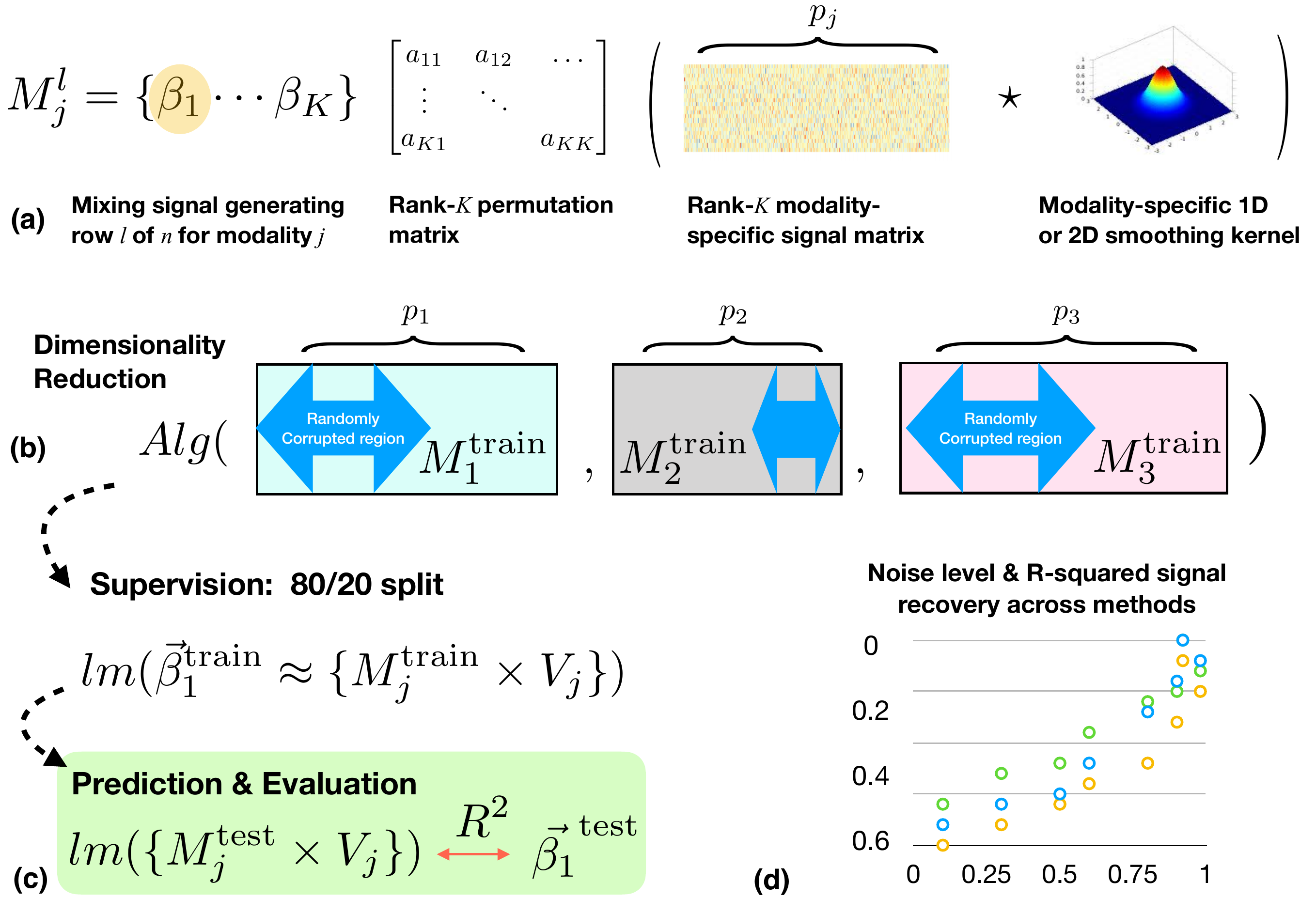} \caption{Conceptual overview of the SiMLR simulation study which defines an inverse problem for signal recovery with known ground truth.  (a) shows the generation of a given row, $M_j^l$, in a single simulated matrix, $M_j$.  The $K$-rank generating basis set is smoothed to induce signal-specific covariation as is present in many types of real biological data.  The highlighted beta is the common signal that we seek to recover. (b) The study also randomly corrupts each of the three generated matrices by eliminating any true signal in some fraction of the matrix.  The fraction of corruption is drawn from a uniform distribution between 0.1 and 0.9 (i.e. 10 and 90 percent corruption). (c)  We define a random 80/20 split for each simulation and learn from the 80 percent of training data.  RGCCA, SGCCA and SiMLR are each run in the dimensionality reduction step.  A linear regression method takes the low-dimensional embeddings, the ground truth signal in the training data and then predicts the test data signal. (d) This process enables us to evaluate the signal recovery performance and how it is impacted by the corruption process.}\label{fig:fig2}
\end{figure}

\hypertarget{simulation-data-and-comparison-to-rgcca-sgcca}{%
\subsection{Simulation data and comparison to RGCCA,
SGCCA}\label{simulation-data-and-comparison-to-rgcca-sgcca}}

SiMLR seeks to solve a multiple modality version of the cocktail party
problem\textsuperscript{43} where the hidden source signals are
distributed across each modality. Therefore, SiMLR assumes that this
common \emph{latent signal} exists across modalities and may be found by
linear projections into a low-dimensional space. We construct simulated
data that matches this setting by constructing 3 matrices from different
(modality-specific) multivariate distributions. Each matrix contains a
common low-dimensional basis (the true latent signal), as illustrated in
Figure 2, which can be recovered by joint dimensionality reduction.
Matrices are generated by the following steps.

\begin{itemize}
\item
  Generate a rank-\(K\) basis set (\(S^K_j\) of size \(K \times p_j\)
  where \(j \in 1, 2, 3\)) of gaussian distributed signal that is
  smoothed by a different amount over each simulation run; \(K\) and
  \(p_j\) vary over simulations.
\item
  Generate ground truth latent signal matrix
  \(B=[\vec{\beta}_1,\cdots,\vec{\beta}_K]\) with \(n\) rows that will
  weight each basis matrix \(S^K_j\). \(B\) is consistent across all
  modalities but \(n\) varies across simulations.
\item
  Generate each \(n \times p_j\) data matrix by computing
  \(M_j~=~B~S^K_j\).
\item
  Replace a percentage of the columns of each matrix \(M_j\) with random
  noise.
\item
  Split the data into 80\% train and 20\% test and run the candidate
  algorithms on each \(M_j^\text{train}\) matrix. Lastly, use linear
  regression to relate the learned embeddings to the true source signal
  (\(\vec{\beta}_1^\text{train}\)) and predict
  \(\vec{\beta}_1^\text{test}\) in the test data from the learned
  embeddings.
\end{itemize}

\noindent The above steps produce data where each of the three 3
matrices is generated from very different distributions but that contain
a common latent signal. The key is that the latent signal is at least
partially consistent (\(\vec{\beta}_1\)) and can, in some cases, be
recovered by joint analysis. Recoverability varies across simulations
due to both corruption and the intrinsic variability of the underlying
generating distributions.

The evaluation criterion then compares the ability of SiMLR to recapture
the known basis with respect to: (a) regularized generalized canonical
correlation analysis (RGCCA); (b) sparse generalized canonical
correlation analysis (SGCCA); (c) permutations of the original data. The
primary evaluation criterion -- accuracy in predicting the true latent
signal -- exhibits that SiMLR's use of cross-modality information and
regularization drives the solution closer to the ground truth basis in
comparison to the other methods. Secondarily, we demonstrate improved
robustness to data corruption. The permutation comparison contrasts the
SiMLR solution to that which would be found when no covariation across
modalities exists.

\hypertarget{signal-recovery}{%
\subsubsection{Signal recovery}\label{signal-recovery}}

An overview of the results is in Figure 3. For each experiment, we run
120 simulations and evaluate the quality of the recovered signal by
training a linear regression algorithm to relate the \emph{learned}
basis to the \emph{true} basis. We then predict the latent signal in
held-out test data (80 percent of subjects are used for training and 20
percent for testing). In this scenario, better performing methods will
lead to more accurate predictions of the latent signal in the tesing
subjects. We can evaluate, by paired \(t\)-test on the recovery
(measured by \(R^2\) of the fit), whether SiMLR performs better than,
equal to or worse than other methods.

The experiments are available in the \texttt{codeocean} capsule via
source \texttt{simulationStudyRGCCA.Rmd} and with recovery outputs
detailed below for each experiment:

\begin{itemize}
\item
  \href{https://files.codeocean.com/files/verified/d885a8e6-0ee8-416a-a58e-88101db23480_v2.0/results.562f0e9b-84c7-4a4a-b0f8-ec2732899be7/simulationStudyRGCCA_CCA_mix_ICA.html}{simulationStudyRGCCA\_CCA\_mix\_ICA.html}:
  The CCA-like objective function with ICA for source separation
  (SiMLR-CCA-ICA). Mean signal recovery \(R^2\) is 0.5043659 with paired
  \(t\)-test relative to RGCCA showing relative improvement at
  \(t=13.27, p\)-value \(< 2.2e-16\) and paired \(t\)-test relative to
  SGCCA showing \(t = 5.0594, p\)-value \(= 1.546e-06\).
\item
  \href{https://files.codeocean.com/files/verified/d885a8e6-0ee8-416a-a58e-88101db23480_v2.0/results.562f0e9b-84c7-4a4a-b0f8-ec2732899be7/simulationStudyRGCCA_CCA_mix_SVD.html}{simulationStudyRGCCA\_CCA\_mix\_SVD.html}:
  The CCA-like objective function with SVD for source separation
  (SiMLR-CCA-SVD). Mean signal recovery \(R^2\) is 0.51233675 with
  paired \(t\)-test relative to RGCCA showing relative improvement at
  \(12.746, p\)-value \(< 2.2e-16\) and paired \(t\)-test relative to
  SGCCA showing \(t = 5.4298, p\)-value \(= 3.028e-07\).
\item
  \href{https://files.codeocean.com/files/verified/d885a8e6-0ee8-416a-a58e-88101db23480_v2.0/results.562f0e9b-84c7-4a4a-b0f8-ec2732899be7/simulationStudyRGCCA_Reg_mix_ICA.html}{simulationStudyRGCCA\_Reg\_mix\_ICA.html}:
  The regression (reconstruction) objective function with ICA
  (SiMLR-Reg-ICA). Mean signal recovery \(R^2\) is 0.49118147 with
  paired \(t\)-test relative to RGCCA showing relative improvement at
  \(t = 12.008, p\)-value \(< 2.2e-16\) and paired \(t\)-test relative
  to SGCCA showing \(t = 3.5184, p\)-value \(= 0.0006158\).
\item
  \href{https://files.codeocean.com/files/verified/d885a8e6-0ee8-416a-a58e-88101db23480_v2.0/results.562f0e9b-84c7-4a4a-b0f8-ec2732899be7/simulationStudyRGCCA_Reg_mix_SVD.html}{simulationStudyRGCCA\_Reg\_mix\_SVD.html}:
  The regression (reconstruction) objective function with SVD
  (SiMLR-Reg-SVD). Mean signal recovery \(R^2\) is 0.49395552 with
  paired \(t\)-test relative to RGCCA showing relative improvement at
  \(t = 11.99, p\)-value \(< 2.2e-16\) and paired \(t\)-test relative to
  SGCCA showing \(t = 3.7482, p\)-value \(= 0.0002763\).
\end{itemize}

The best overall was SiMLR-CCA-SVD with an average \(R^2\) recovery of
0.51 while SGCCA and RGCCA score 0.45 and 0.35 respectively.

\hypertarget{sensitivity-to-amount-of-corrupted-data}{%
\subsubsection{Sensitivity to amount of corrupted
data}\label{sensitivity-to-amount-of-corrupted-data}}

As above, for each of the 120 simulations, a varying degree of
corruption to each matrix is performed. That is, a random percentage of
the matrix that contains true signal is replaced with noise signal with
no relationship to the latent ground truth. The amount of corruption
varies between 10 and 90 percent of the column entries. This enables us
to test the degree to which recovery performance can be predicted from
the amount of corruption where corruption is represented as a 3-vector
for each experiment where each entry in the vector codifies the amount
of corruption. RGCCA performance (\(R^2\)) is related to corruption with
\(p\)-value 0.01179. SGCCA performance is related with \(p\)-value
0.0001525. SiMLR-CCA-ICA is related with \(p\)-value 0.01045.
SiMLR-CCA-SVD is related with \(p\)-value 0.04103. SiMLR-Reg-ICA is
related with \(p\)-value 0.0223. SiMLR-Reg-SVD is related with
\(p\)-value 0.006516. As such, SGCCA (in this experiment) is most
sensitive of these methods to corrupted data and SiMLR-CCA-SVD is least
so. Inspection of the \(R^2\) performance plots indicates that the
impact of corruption is not insubstantial with 24 of 120 RGCCA
experiments leading to \(R^2\) less than 0.2. For SGCCA, SiMLR-CCA-ICA,
SiMLR-CCA-SVD, SiMLR-Reg-ICA and SiMLR-Reg-SVD, these totals are 13, 6,
5, 3 and 4 respectively. From this perspective of a performance cutoff,
SGCCA does slightly better than RGCCA. However, both methods still are
less reliable than all variants of SiMLR. In the remaining evaluation
studies, we focus on contrasting SGCCA with SiMLR because they both
involve feature selection which is more appropriate for the \(p >> n\)
cases that we investigate.

\hypertarget{recovery-of-two-latent-signals}{%
\subsubsection{Recovery of two latent
signals}\label{recovery-of-two-latent-signals}}

A novel hallmark of SiMLR's design is that it relies on source
separation to define the intermediate latent space used during the
optimization of its feature matrices. This suggests that SiMLR should
exhibit improved recovery of not just a single latent vector space (as
above) but several at once. We demonstrate this advantage in an
experiment that follows the same design as above but evaluates recovery
of two latent signals. As expected, SiMLR outperforms SGCCA to an even
greater degree in this more challenging scenario where, for the second
signal to be recovered, SiMLR outperforms SGCCA as measured by paired
\(t\)-test (239 degrees of freedom, t = 6.41, \(t\)-value = 7.55e-10).
The details of these results are in supplementary information and the
experimental code is here (link):
\url{http://github.com/stnava/symlrExamples/blob/master/figures/simulationStudyRGCCA_2components.pdf}.

\newpage

\begin{figure}
\includegraphics[width=1\linewidth]{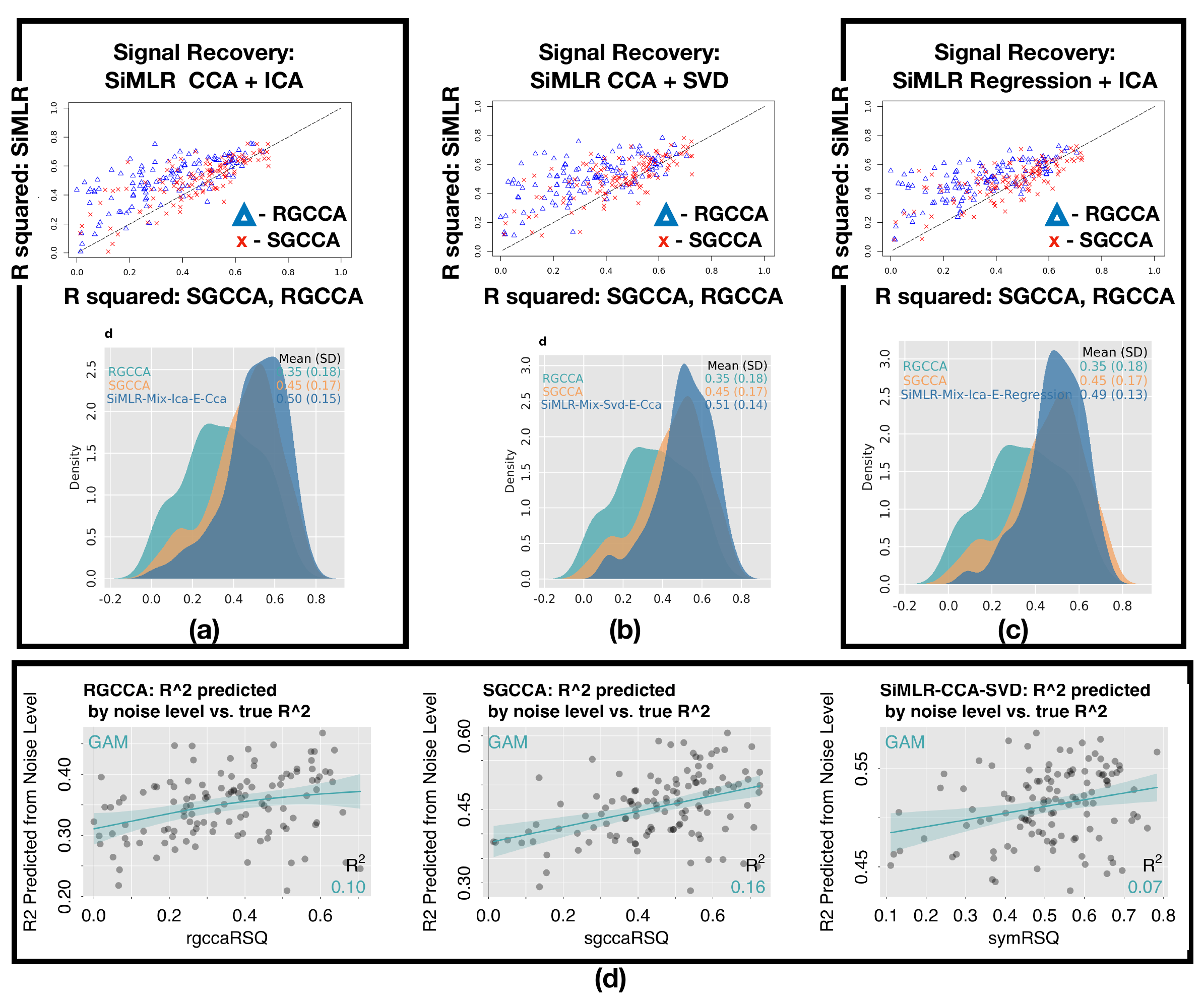} \caption{SiMLR simulation study results: sensitivity to noise and ability to recover signal.  In each panel, (a-c), the SiMLR signal recovery performance in terms of $R^2$ is plotted against RGCCA and SGCCA performance.  Thus, higher scores are better and points above the diagonal dotted line show superior SiMLR performance in pair-wise fashion. (a) Demonstrates performance of signal recovery of SiMLR with the CCA energy and ICA source separation method.  (b) Demonstrates performance of signal recovery of SiMLR with the CCA energy and SVD source separation method. (c) Demonstrates performance of signal recovery of SiMLR with the regression energy and ICA source separation method.  The lower plots in (d) show how performance is impacted by the amount of matrix corruption.  Here, lower scores are better.  In this simulation study, SiMLR with CCA and SVD source separation does best in terms of both raw scores and pairwise test statistics.}\label{fig:fig3}
\end{figure}

\newpage

\hypertarget{cancer-survival-prediction-from-multi-omic-data}{%
\subsection{Cancer survival prediction from multi-omic
data}\label{cancer-survival-prediction-from-multi-omic-data}}

We compare dimensionality reduction methods with respect to glioblastoma
(GBM) survival prediction based on multi-omics data. The biological data
includes \(n=208\) subjects with:

\begin{itemize}
\item
  Gene expression: with \(p=20,531\) predictors; see\textsuperscript{44}
  for more information on this modality;
\item
  Methylomics/DNA methylation: with \(p=5,000\) predictors;
  see\textsuperscript{45} for background;
\item
  Transcriptomics/micro RNA expression: with \(p=1,046\) predictors;
  see\textsuperscript{46} for background;
\end{itemize}

and is complemented by survival data (number of days since diagnosis and
whether or not death has occurred at that time). Additional cohort
information, such as age and gender, is also available but was not used
here
\footnote{We have results from experiments that include more demographic data along with multiomics to predict survival in not only GBM but also glioblastoma and melanoma. This may be shared publicly in the future after more validation of the data organization. Please contact the authors if interested.}

The hypothesis is that this collection of measurements, which track the
biological/genetic dynamics of tumor activity, will improve prediction
of patient-specific outcomes. However, these data are fairly
high-dimensional relative to the number of subjects. As such, targeted
dimensionality reduction is needed to overcome the \(p >> n\) problem
(where \(p\), here, refers to predictors) in order to allow
low-dimensional versions of these predictors (i.e.~embeddings) to be
used in a classical regression context.

The model is a Cox proportional hazards regression
model\textsuperscript{47} implemented in the \texttt{coxph} function in
the \texttt{survival} package\textsuperscript{48}. We evaluate
concordance in test data via the \texttt{survcomp} package. Concordance
is similar to a rank correlation method and is used to assess agreement
of the predicted outcomes (chance of death, given survival time and the
multi-omic predictors) with true outcomes. Its value under the null
hypothesis of no predictive value is 0.5. Values greater than roughly
0.6 show some evidence of predictive power in this application
context\textsuperscript{49,50}. Data is the GBM set from
\href{http://acgt.cs.tau.ac.il/multi_omic_benchmark/download.html}{the
multi-omic benchmark collection}. The published paper associated with
this benchmark is\textsuperscript{51}. We chose the GBM data as it
allowed a train-test split with sufficient variability of survival with
respect to the \(n\) in both train and test groups, in particular in the
latter. GBM shows among the highest differential survival among clusters
reported in\textsuperscript{51}. These data are collected from The
Cancer Genome Atlas (TCGA).

The benchmark paper above showed that ``with respect to survival, MCCA
had the total best prognostic value'' where MCCA refers to multiple
canonical correlation analysis ( pairwise CCA across all pairs
)\textsuperscript{52}. Thus, the comparison between SGCCA and SiMLR is
pertinent. Nevertheless, these approaches should not be considered as
the best strategy given that single-omic analysis did nearly as well.
The authors find that ``analysis of multi-omics data does not
consistently provide better prognostic value and clinical significance
compared to analysis of single-omic data alone, especially when
different single-omics are used for each cancer types''.

The study design is fairly simple. For both SGCCA and SiMLR and over 50
runs, we:

\begin{itemize}
\item
  split the data into 80\% training and 20\% testing sets.
\item
  in training data, we perform supervised dimensionality reduction where
  the 'omics data is jointly reduced with both death and survival time
  acting as a fourth matrix;
\item
  train a Cox model with the low-dimensional bases derived from the
  'omics data in the previous step;
\item
  predict the survival outcome in test data and evaluate accuracy with
  the concordance metric.
\end{itemize}

Under this design, a better method will both produce higher concordance
values and produce more concordance values that meet or exceed the value
of 0.6 which is considered a threshold of moderate
agreement\textsuperscript{53} and has been reported in recent studies
for reasonably performing methods\textsuperscript{49,50}. We repeat the
above experiments over 50 splits of the data in order to gain an
empirical estimate of the difference in performance between SiMLR and
SGCCA with different input data. We also test at three different
sparseness levels thereby comparing the performance of solutions that
yield feature vectors with low (25\%), moderate (50\%) and high (75\%)
sparseness.

In this evaluation, SiMLR with the reconstruction/regression energy
shows an advantage over SGCCA in terms of predictive performance as
measured by concordance in test data. Average concordance for SiMLR plus
reconstruction is 0.64. The covariance energy and SGCCA perform nearly
identically and, on average, do not exceed 0.6 concordance. Furthermore,
in contrast to SGCCA, SiMLR's feature vectors are not only sparse but
also smooth and unsigned (non-negative) which aids interpretation and
may prevent overfitting, thus improving generalization. In the example
code at codeocean \texttt{simlr\_TCGA\_survival.Rmd}, graph-based
regularization parameters are selected to include 2.5\% of the
predictors in each predictor 'omics matrix (see the call to the
\texttt{regularizeSimlr} function). As such, regularization is present
but neither overwhelming nor optimized for this data. I.e. this value
was chosen based on the desire for a small amount of denoising in the
solution space. Neither method was optimized for this problem in terms
of data selection, parameter or pre-processing choices. Moreover, the
authors are not domain experts in this field. As such, this acts as a
fairly unbiased comparison of these tools.

\newpage

\begin{figure}
\includegraphics[width=1\linewidth]{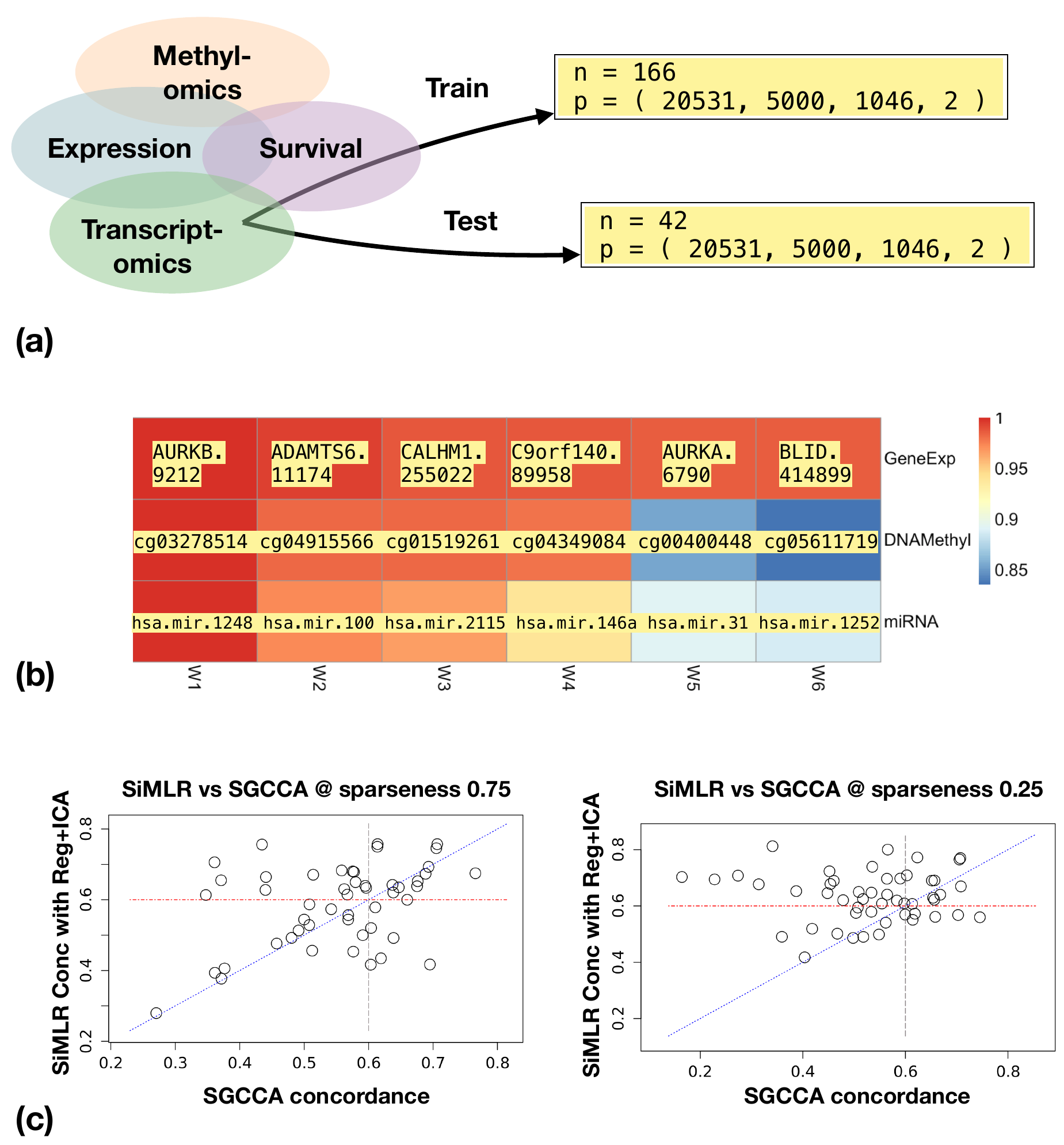} \caption{Multi-omic study: supervised cancer survival prediction comparison to SGCCA.  In (a), we show the train-test split numbers for each of the simulation runs as well as the $n$ and $p$ for each measurement type: gene expression, methylomics, transcriptomics and survival data. (b) Shows the most highly weighted (top 6) predictors selected for each of the modalities where the weights are normalized such that the maximum value is one.  Because the feature vectors are non-negative, these lead to interpretable regression weights when the embeddings are combined with the Cox proportional-hazards model.  (c) Shows plots of SiMLR concordance versus SGCCA concordance over all 50 runs.  As such, points above the red dotted line and above the blue dotted line show where SiMLR performs reasonably and better than SGCCA. At left in (c) shows a case where SiMLR and SGCCA demonstrate close performance and (right) where SiMLR outperforms SGCCA the most.}\label{fig:fig4}
\end{figure}

Figure 4 summarizes overall performance differences, along with the full
output available on the \texttt{codeocean} platform. A brief summary of
results is here for each comparison with pointers to the appropriate
computational output file:

\begin{itemize}
\tightlist
\item
  \href{simlr_TCGA_survival_CCA_mix_ICA.html}{simlr\_TCGA\_survival\_CCA\_mix\_ICA}:
  SiMLR-CCA-ICA is compared to SGCCA at the three sparseness levels. At
  sparseness:

  \begin{itemize}
  \tightlist
  \item
    0.25, a paired \(t\)-test shows \(t = -1.04\), \(p\)-value = 0.30
    (where positive \(t\)-statistic indicates better performance by
    SiMLR);
  \item
    0.50, a paired \(t\)-test shows \(t = 1.01\), \(p\)-value = 0.32.
  \item
    0.75, a paired \(t\)-test shows \(t = -1.41\), \(p\)-value = 0.16.
  \end{itemize}
\item
  \href{simlr_TCGA_survival_CCA_mix_SVD.html}{simlr\_TCGA\_survival\_CCA\_mix\_SVD}:
  SiMLR-CCA-SVD comparisons show:

  \begin{itemize}
  \tightlist
  \item
    at sparseness 0.25, \(t\) = 0.72, \(p\)-value = 0.48;
  \item
    at sparseness 0.50, \(t\) = 0.09, \(p\)-value = 0.93;
  \item
    at sparseness 0.75, \(t\) = -0.82, \(p\)-value = 0.41.
  \end{itemize}
\item
  \href{simlr_TCGA_survival_Reg_mix_ICA.html}{simlr\_TCGA\_survival\_Reg\_mix\_ICA}:
  SiMLR-Reg-ICA comparisons show:

  \begin{itemize}
  \tightlist
  \item
    at sparseness 0.25, \(t\) = 4.53, \(p\)-value = 3.74e-05;
  \item
    at sparseness 0.50, \(t\) = 4.46, \(p\)-value = 4.85e-05;
  \item
    at sparseness 0.75, \(t\) = 1.81, \(p\)-value = 0.08.
  \end{itemize}
\item
  \href{simlr_TCGA_survival_Reg_mix_SVD.html}{simlr\_TCGA\_survival\_Reg\_mix\_SVD}:
  SiMLR-Reg-SVD comparisons show:

  \begin{itemize}
  \tightlist
  \item
    at sparseness 0.25, \(t\) = 4.14, \(p\)-value = 0.0001;
  \item
    at sparseness 0.50, \(t\) = 3.61, \(p\)-value = 0.0007;
  \item
    at sparseness 0.75, \(t\) = 2.55, \(p\)-value = 0.0141.
  \end{itemize}
\end{itemize}

In summary, SiMLR with reconstruction performs statistically
equivalently or better on average than SGCCA in this problem with
SiMLR-Reg-ICA showing the best results over all sparseness values.
SiMLR-CCA performed equivalently to SGCCA. In addition, as shown in
Figure 4 (b), SiMLR provides feature weights that have a consistent sign
and that allow the embeddings to be treated as weighted averages. This
means that, for example, high values for gene expression-related
embeddings are due to the contributions of individual genes that exhibit
high expression levels. This aids interpretation of statistical models
based on these embeddings because both the units and the directionality
of these derived variables are clear.

\newpage

\hypertarget{simlr-and-sgcca-applied-to-ptbp-brain-age}{%
\subsection{SiMLR and SGCCA applied to PTBP brain
age}\label{simlr-and-sgcca-applied-to-ptbp-brain-age}}

The pediatric template of brain perfusion (PTBP\textsuperscript{54})
includes freely available multiple modality neuroimaging consistently
collected in a cohort of subjects between ages 7 and 18 years of age.
PTBP also includes a variety of demographic and cognitive measurements.
A relevant reference analysis of this data is available
in\textsuperscript{55}.

We provide pre-processed (machine learning ready) matrix format for
three measurements taken in 97 subjects: voxelwise cortical
thickness\textsuperscript{56}, fractional anisotropy (FA) derived from
diffusion tensor imaging and cerebral blood flow (CBF) all at the
voxel-wise level at 1mm resolution. The dimensionality of the matrices
are 97 \(\times\) 515,317 for thickness and CBF and 97 \(\times\)
438,394 for FA. The development-related phenotype matrix consists of the
subjects' sex, chronological age, total IQ score, verbal IQ score and
performance IQ score. The IQ variables are highly correlated.

Figure 5 summarizes the design of the study. First, a 5-fold
cross-validation grouping of subjects is defined. For each fold, SiMLR
and SGCCA are run with parameters that are set to select interpretable
``network''-like components. In this example, we choose these parameters
specifically at higher sparseness levels to facilitate interpretability.
We record computation time as well as the embedding vectors for each
modality. We then train, within each fold, a linear regression model to
predict age and IQ-related variables from the neuroimaging embeddings.
This is a form of principal component regression. These predictions are
stored for each fold to facilitate a final comparison of performance
across all folds. We use this technique, rather than repeated resampling
as in prior studies, in part because the run-time for this problem can
be relatively long, up to 235 minutes for SGCCA. The following results
are drawn from codeocean \url{simlr_PTBP_supervised_ICA} and
\url{simlr_PTBP_supervised_SVD}. The ICA and SVD differentiate which
source separation method is used. Both show the RGCCA results and a
summary of all cross-validated prediction is in the csv files called
\texttt{simlr\_ptbp\_summary\_results\_ica} or
\texttt{simlr\_ptbp\_summary\_results\_svd}.

\hypertarget{computation-time}{%
\subsubsection{Computation time}\label{computation-time}}

In this example, SGCCA and SiMLR run, on the \texttt{codeocean}
platform, demonstrate overall similar run-time with a few exceptions.
These exceptions are caused by data-dependent longer convergence times.
SGCCA runs, over each of five folds, for 235, 29, 29, 30 and 33 minutes.
SiMLR with CCA and ICA runs for 105, 46, 46, 41 and 46 minutes. SiMLR
with CCA and SVD runs for 78, 58, 94, 53 and 56 minutes. SiMLR with
regression and ICA runs for 44, 47, 55, 39 and 60 minutes. SiMLR with
regression and SVD runs for 28, 31, 41, 54 and 33 minutes. Overall
differences in run-time likely depend on convergence settings as well as
the variability of the energy function combined with the input data.

\hypertarget{prediction-outcomes}{%
\subsubsection{Prediction outcomes}\label{prediction-outcomes}}

Figure 5 demonstrates the predictions' mean absolute error (MAE) for
each algorithm that we tested:

\begin{itemize}
\tightlist
\item
  SGCCA yields 2.0 years MAE;
\item
  SiMLR-Reg-SVD yields 1.70 years;
\item
  SiMLR-CCA-ICA yields 1.64 years;
\item
  SiMLR-Reg-ICA yields 1.58 years;
\item
  SiMLR-CCA-SVD yields 1.44 years.
\end{itemize}

\noindent None of the methods perform well for predicting IQ-related
scores. However, both SiMLR and SGCCA component regression produce
reasonable predictions of brain age\textsuperscript{57}. These values
reported here are competitive with those reported
in\textsuperscript{55}. The MAE differences translate to a statistically
significant improvement in performance between SiMLR (all variants) and
SGCCA (best result \(p\)-value = 0.0002965, worst \(p\)-value = 0.04692
). At the individual prediction level, this means that SiMLR produces a
more accurate age in 61 of 97 cases for SiMLR-CCA-SVD.

\newpage

\begin{figure}
\includegraphics[width=1\linewidth]{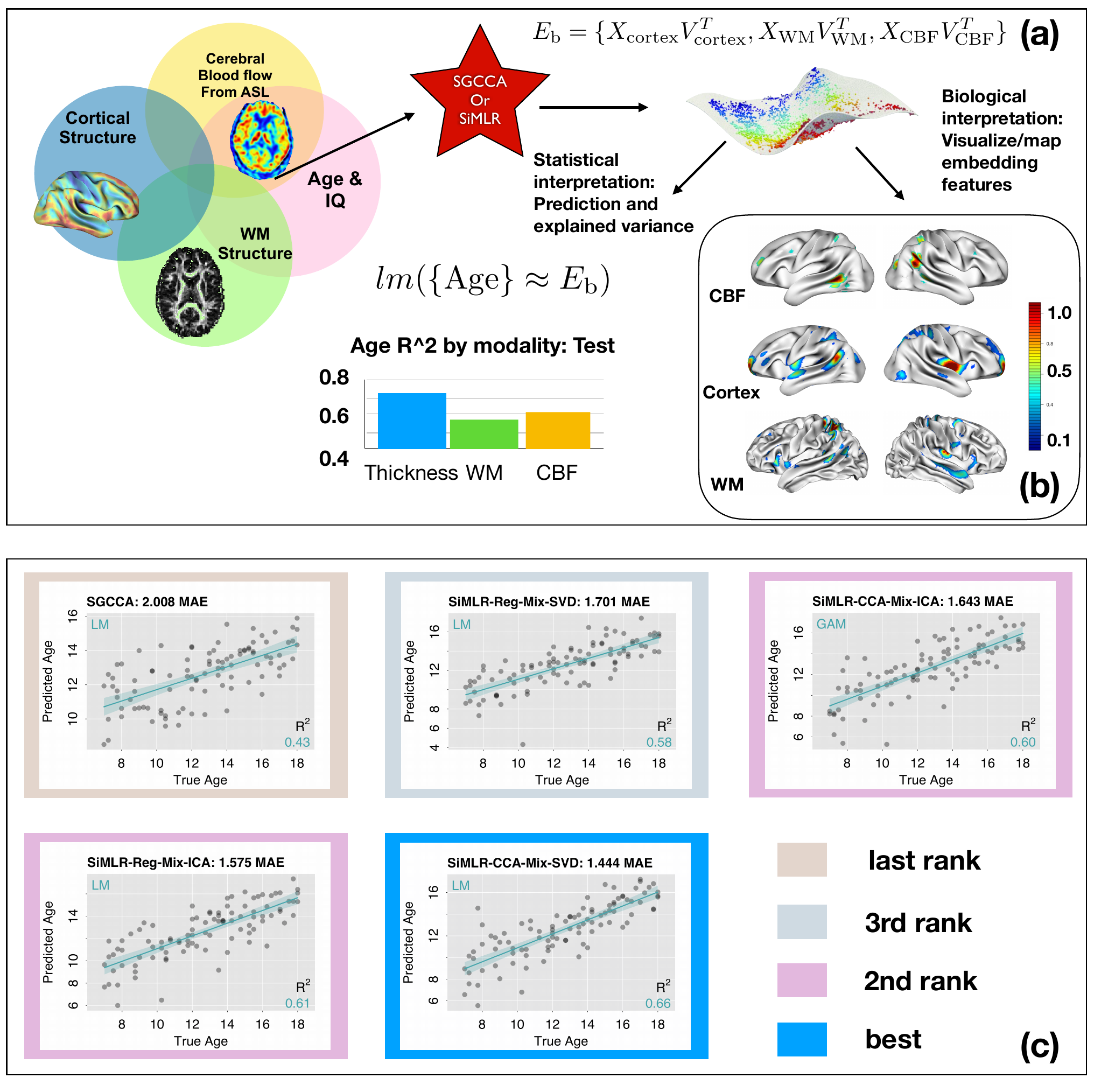} \caption{PTBP fully supervised brain age prediction: comparison to SGCCA.  Brain age is the subject's age predicted from neuroimaging data.  Four matrices are input where cortical thickness, white matter integrity and cerebral blood flow derive from different types of neuroimages; the fourth modality describes brain maturation in terms of age and IQ measurements. Panel (a) shows the overall study design where embeddings are computed as in prior examples and then passed downstream to facilitate statistical and biological interpretation.  The first phase of statistical interpretation ( the bar plots ) compares the ability of each modality to predict age independently and suggests thickness is most predictive (when acting alone); WM and CBF have close performance to each other (data drawn from best performing method).  In (b), we show the feature vectors from the best performing method noting that the weights are relative to each feature vector where its values are scaled to zero to one.  In (c) we show the ability to predict real age from the brain where SiMLR-CCA-SVD does best; none of the methods perform well for IQ prediction.  Standard error bars are shown as gray shaded regions around a best-fit line. $R^2$ for the predicted model fit is also shown.}\label{fig:fig5}
\end{figure}

\newpage

\hypertarget{simlr-and-sgcca-applied-to-imaging-genetics-data}{%
\subsection{SiMLR and SGCCA applied to imaging-genetics
data}\label{simlr-and-sgcca-applied-to-imaging-genetics-data}}

Pediatric Imaging, Neurocognition, and Genetics (PING)
data\textsuperscript{58} offers the opportunity to jointly study two
types of neuroimaging, anxiety and depression related
SNPs\textsuperscript{2} and self-reported scores of anxiety and
depression. The training portion of the data is defined by subjects who
have only neuroimaging and SNPs. This allows us to perform
dimensionality reduction in training subjects alone (\(n\)=508) to
identify a much lower dimensional space that encodes the variability
induced jointly by SNPs and brain structure. The test set is
distinguished by individuals who have not only imaging and genetics
measurements but also self-reported measures of anxiety and depression.
We perform inference in the test set (\(n\)=162) to determine which, if
any, of the learned embeddings relate to these scores.

The evaluation criterion, here, is inferential i.e.~we prefer the method
that leads to embeddings with greater relationship to the clinical
scores. This exploratory study is shared in supplementary information.
For this reason, we keep discussion, here, brief. Primarily, SiMLR
identifies more signal related to anxiety and depression in the
inferential portion of the study, when compared to SGCCA. I.e. more
components relate to self-report anxiety and depression scores -- with
both SNPs and brain structure (thickness and white matter integrity,
like PTBP) contributing -- when using SiMLR compared to SGCCA. SiMLR
leads to 3 components whereas SGCCA only identifies a single component
related to anxiety. More importantly, however, we noted a severe
difference in computation time. This study computes 40 components from
high-dimensional data. SGCCA takes over 24 hours to compute these
components. SiMLR (all variants) takes less than an hour. The primary
difference between this study and the others included as examples is
that the number of rows and the number of columns is relatively large
(for training, \(n\)=508 and
\(p_\text{thickness}=66,565, p_2=68,966, p_3=4,309\)). As such, the
advantage SGCCA gains by working in the dual space may be overwhelmed by
the combined cost of relatively large covariance matrices and the need
to perform deflation for each set of components. In contrast, SiMLR
computes the feature matrix for each modality in one pass through the
optimization.

\noindent **NOTE: REVIEWERS** the only change here is a brief paragraph
describing a new supplementary result.

We also provide a related supplementary result in multi-omic Alzheimer's
disease neuroimaging initiative (ADNI) data. This result shows another
way to relate imaging and cognition to genetic measurements: through
polygenic risk measurements. Polygenic risk scores effectively reduce
the dimensionality of genetic data based on an \emph{a priori} weighted
sum of trait-associated alleles. The supplementary document (entitled
``SiMLR multi-omics study in ADNI'') contrasts SiMLR, RGCCA and SGCCA
applied to tabular data where \(n >> p\). The results of this joint
reduction repeat trends shown elsewhere in this document; however, the
difference between sparse and unconstrained dimensionality reduction is
relatively less due to the more classical setting (\(n >> p\)). This
demonstrates that SiMLR can be used effectively, like RGCCA, even when a
dataset is already relatively well-powered.

\newpage

  \providecommand{\huxb}[2]{\arrayrulecolor[RGB]{#1}\global\arrayrulewidth=#2pt}
  \providecommand{\huxvb}[2]{\color[RGB]{#1}\vrule width #2pt}
  \providecommand{\huxtpad}[1]{\rule{0pt}{#1}}
  \providecommand{\huxbpad}[1]{\rule[-#1]{0pt}{#1}}

\begin{table}[ht]
\begin{centerbox}
\begin{threeparttable}
\captionsetup{justification=centering,singlelinecheck=off}
\caption{Summary of RGCCA-SGCCA-SiMLR comparison results. The $x+/-y$ indicates mean and standard deviation values in the table below. RGCCA = regularized generalized canonical correlation analysis; SGCCA = sparse generalized canonical correlation analysis; Sim = similarity-driven multivariate linear reconstruction (SiMLR); Reg = regression; CCA = absolute canonical covariance; ICA = ICA blind source separation (BSS) method; SVD = SVD (BSS) method.  Best results are highlighted in cadet blue; worst in antiquewhite. The second ranking approach is in pink. SiMLR with the absolute canonical covariance similarity measurement and SVD (SimCCASVD) as a BSS method performs best overall.   For the multi-omics example, SiMLR with the regression energy and ICA BSS method (SimRegICA) outperforms SGCCA most consistently across sparseness levels, provides closely competitive performance overall, and is highlighted in pink.  In brainAge, all SiMLR variants perform substantially better than SGCCA.  The PING examples are exploratory analyses described in the supplementary information as we cannot directly share the data. The "n comp" description in the PING table refers to the number of significant components related to either anxiety or depression.  The last row summarizes our overall ranking and explains the relationship of cell color to the ranking system.  The rank is calculated by counting instances of a given rank across columns.}
 \label{tab:table1}
\setlength{\tabcolsep}{0pt}
\begin{tabular}{l l l l l l l l}

\hhline{}
\arrayrulecolor{black}

\multicolumn{1}{!{\huxvb{0, 0, 0}{0}}l!{\huxvb{0, 0, 0}{0}}}{\huxtpad{6pt + 1em}\raggedright \hspace{6pt} \textbf{{\fontsize{8pt}{9.6pt}\selectfont study}} \hspace{6pt}\huxbpad{6pt}} &
\multicolumn{1}{l!{\huxvb{0, 0, 0}{0}}}{\huxtpad{6pt + 1em}\raggedright \hspace{6pt} \textbf{{\fontsize{8pt}{9.6pt}\selectfont RGCCA}} \hspace{6pt}\huxbpad{6pt}} &
\multicolumn{1}{l!{\huxvb{0, 0, 0}{0}}}{\huxtpad{6pt + 1em}\raggedright \hspace{6pt} \textbf{{\fontsize{8pt}{9.6pt}\selectfont SGCCA}} \hspace{6pt}\huxbpad{6pt}} &
\multicolumn{1}{l!{\huxvb{0, 0, 0}{0}}}{\huxtpad{6pt + 1em}\raggedright \hspace{6pt} \textbf{{\fontsize{8pt}{9.6pt}\selectfont SimCCAICA}} \hspace{6pt}\huxbpad{6pt}} &
\multicolumn{1}{l!{\huxvb{0, 0, 0}{0}}}{\huxtpad{6pt + 1em}\raggedright \hspace{6pt} \textbf{{\fontsize{8pt}{9.6pt}\selectfont SimCCASVD}} \hspace{6pt}\huxbpad{6pt}} &
\multicolumn{1}{l!{\huxvb{0, 0, 0}{0}}}{\huxtpad{6pt + 1em}\raggedright \hspace{6pt} \textbf{{\fontsize{8pt}{9.6pt}\selectfont SimRegICA}} \hspace{6pt}\huxbpad{6pt}} &
\multicolumn{1}{l!{\huxvb{0, 0, 0}{0}}}{\huxtpad{6pt + 1em}\raggedright \hspace{6pt} \textbf{{\fontsize{8pt}{9.6pt}\selectfont SimRegSVD}} \hspace{6pt}\huxbpad{6pt}} &
\multicolumn{1}{l!{\huxvb{0, 0, 0}{0}}}{\huxtpad{6pt + 1em}\raggedright \hspace{6pt} \textbf{{\fontsize{8pt}{9.6pt}\selectfont metric}} \hspace{6pt}\huxbpad{6pt}} \tabularnewline[-0.5pt]

\hhline{>{\huxb{0, 0, 0}{0.4}}->{\huxb{0, 0, 0}{0.4}}->{\huxb{0, 0, 0}{0.4}}->{\huxb{0, 0, 0}{0.4}}->{\huxb{0, 0, 0}{0.4}}->{\huxb{0, 0, 0}{0.4}}->{\huxb{0, 0, 0}{0.4}}->{\huxb{0, 0, 0}{0.4}}-}
\arrayrulecolor{black}

\multicolumn{1}{!{\huxvb{0, 0, 0}{0}}l!{\huxvb{0, 0, 0}{0}}}{\cellcolor[RGB]{242, 242, 242}\huxtpad{6pt + 1em}\raggedright \hspace{6pt} {\fontsize{8pt}{9.6pt}\selectfont Signal-Sens.} \hspace{6pt}\huxbpad{6pt}} &
\multicolumn{1}{l!{\huxvb{0, 0, 0}{0}}}{\cellcolor[RGB]{250, 235, 215}\huxtpad{6pt + 1em}\raggedright \hspace{6pt} {\fontsize{8pt}{9.6pt}\selectfont 0.35+/-0.18} \hspace{6pt}\huxbpad{6pt}} &
\multicolumn{1}{l!{\huxvb{0, 0, 0}{0}}}{\cellcolor[RGB]{242, 242, 242}\huxtpad{6pt + 1em}\raggedright \hspace{6pt} {\fontsize{8pt}{9.6pt}\selectfont 0.45+/-0.17} \hspace{6pt}\huxbpad{6pt}} &
\multicolumn{1}{l!{\huxvb{0, 0, 0}{0}}}{\cellcolor[RGB]{255, 192, 203}\huxtpad{6pt + 1em}\raggedright \hspace{6pt} {\fontsize{8pt}{9.6pt}\selectfont 0.5+/-0.15} \hspace{6pt}\huxbpad{6pt}} &
\multicolumn{1}{l!{\huxvb{0, 0, 0}{0}}}{\cellcolor[RGB]{152, 245, 255}\huxtpad{6pt + 1em}\raggedright \hspace{6pt} {\fontsize{8pt}{9.6pt}\selectfont 0.51+/-0.14} \hspace{6pt}\huxbpad{6pt}} &
\multicolumn{1}{l!{\huxvb{0, 0, 0}{0}}}{\cellcolor[RGB]{242, 242, 242}\huxtpad{6pt + 1em}\raggedright \hspace{6pt} {\fontsize{8pt}{9.6pt}\selectfont 0.49+/-0.13} \hspace{6pt}\huxbpad{6pt}} &
\multicolumn{1}{l!{\huxvb{0, 0, 0}{0}}}{\cellcolor[RGB]{242, 242, 242}\huxtpad{6pt + 1em}\raggedright \hspace{6pt} {\fontsize{8pt}{9.6pt}\selectfont 0.49+/-0.14} \hspace{6pt}\huxbpad{6pt}} &
\multicolumn{1}{l!{\huxvb{0, 0, 0}{0}}}{\cellcolor[RGB]{242, 242, 242}\huxtpad{6pt + 1em}\raggedright \hspace{6pt} {\fontsize{8pt}{9.6pt}\selectfont R-squared} \hspace{6pt}\huxbpad{6pt}} \tabularnewline[-0.5pt]

\hhline{}
\arrayrulecolor{black}

\multicolumn{1}{!{\huxvb{0, 0, 0}{0}}l!{\huxvb{0, 0, 0}{0}}}{\huxtpad{6pt + 1em}\raggedright \hspace{6pt} {\fontsize{8pt}{9.6pt}\selectfont Noise-Sens.} \hspace{6pt}\huxbpad{6pt}} &
\multicolumn{1}{l!{\huxvb{0, 0, 0}{0}}}{\huxtpad{6pt + 1em}\raggedright \hspace{6pt} {\fontsize{8pt}{9.6pt}\selectfont 0.09} \hspace{6pt}\huxbpad{6pt}} &
\multicolumn{1}{l!{\huxvb{0, 0, 0}{0}}}{\cellcolor[RGB]{250, 235, 215}\huxtpad{6pt + 1em}\raggedright \hspace{6pt} {\fontsize{8pt}{9.6pt}\selectfont 0.16} \hspace{6pt}\huxbpad{6pt}} &
\multicolumn{1}{l!{\huxvb{0, 0, 0}{0}}}{\huxtpad{6pt + 1em}\raggedright \hspace{6pt} {\fontsize{8pt}{9.6pt}\selectfont 0.09} \hspace{6pt}\huxbpad{6pt}} &
\multicolumn{1}{l!{\huxvb{0, 0, 0}{0}}}{\cellcolor[RGB]{152, 245, 255}\huxtpad{6pt + 1em}\raggedright \hspace{6pt} {\fontsize{8pt}{9.6pt}\selectfont 0.06} \hspace{6pt}\huxbpad{6pt}} &
\multicolumn{1}{l!{\huxvb{0, 0, 0}{0}}}{\cellcolor[RGB]{255, 192, 203}\huxtpad{6pt + 1em}\raggedright \hspace{6pt} {\fontsize{8pt}{9.6pt}\selectfont 0.07} \hspace{6pt}\huxbpad{6pt}} &
\multicolumn{1}{l!{\huxvb{0, 0, 0}{0}}}{\huxtpad{6pt + 1em}\raggedright \hspace{6pt} {\fontsize{8pt}{9.6pt}\selectfont 0.1} \hspace{6pt}\huxbpad{6pt}} &
\multicolumn{1}{l!{\huxvb{0, 0, 0}{0}}}{\huxtpad{6pt + 1em}\raggedright \hspace{6pt} {\fontsize{8pt}{9.6pt}\selectfont R-squared} \hspace{6pt}\huxbpad{6pt}} \tabularnewline[-0.5pt]

\hhline{}
\arrayrulecolor{black}

\multicolumn{1}{!{\huxvb{0, 0, 0}{0}}l!{\huxvb{0, 0, 0}{0}}}{\cellcolor[RGB]{242, 242, 242}\huxtpad{6pt + 1em}\raggedright \hspace{6pt} {\fontsize{8pt}{9.6pt}\selectfont Multi-omic} \hspace{6pt}\huxbpad{6pt}} &
\multicolumn{1}{l!{\huxvb{0, 0, 0}{0}}}{\cellcolor[RGB]{242, 242, 242}\huxtpad{6pt + 1em}\raggedright \hspace{6pt} {\fontsize{8pt}{9.6pt}\selectfont  N/A } \hspace{6pt}\huxbpad{6pt}} &
\multicolumn{1}{l!{\huxvb{0, 0, 0}{0}}}{\cellcolor[RGB]{242, 242, 242}\huxtpad{6pt + 1em}\raggedright \hspace{6pt} {\fontsize{8pt}{9.6pt}\selectfont 0.56+/-0.12} \hspace{6pt}\huxbpad{6pt}} &
\multicolumn{1}{l!{\huxvb{0, 0, 0}{0}}}{\cellcolor[RGB]{242, 242, 242}\huxtpad{6pt + 1em}\raggedright \hspace{6pt} {\fontsize{8pt}{9.6pt}\selectfont 0.56+/-0.13} \hspace{6pt}\huxbpad{6pt}} &
\multicolumn{1}{l!{\huxvb{0, 0, 0}{0}}}{\cellcolor[RGB]{250, 235, 215}\huxtpad{6pt + 1em}\raggedright \hspace{6pt} {\fontsize{8pt}{9.6pt}\selectfont 0.56+/-0.14} \hspace{6pt}\huxbpad{6pt}} &
\multicolumn{1}{l!{\huxvb{0, 0, 0}{0}}}{\cellcolor[RGB]{152, 245, 255}\huxtpad{6pt + 1em}\raggedright \hspace{6pt} {\fontsize{8pt}{9.6pt}\selectfont 0.64+/-0.08} \hspace{6pt}\huxbpad{6pt}} &
\multicolumn{1}{l!{\huxvb{0, 0, 0}{0}}}{\cellcolor[RGB]{255, 192, 203}\huxtpad{6pt + 1em}\raggedright \hspace{6pt} {\fontsize{8pt}{9.6pt}\selectfont 0.64+/-0.11} \hspace{6pt}\huxbpad{6pt}} &
\multicolumn{1}{l!{\huxvb{0, 0, 0}{0}}}{\cellcolor[RGB]{242, 242, 242}\huxtpad{6pt + 1em}\raggedright \hspace{6pt} {\fontsize{8pt}{9.6pt}\selectfont Concordance} \hspace{6pt}\huxbpad{6pt}} \tabularnewline[-0.5pt]

\hhline{}
\arrayrulecolor{black}

\multicolumn{1}{!{\huxvb{0, 0, 0}{0}}l!{\huxvb{0, 0, 0}{0}}}{\huxtpad{6pt + 1em}\raggedright \hspace{6pt} {\fontsize{8pt}{9.6pt}\selectfont brainAge} \hspace{6pt}\huxbpad{6pt}} &
\multicolumn{1}{l!{\huxvb{0, 0, 0}{0}}}{\huxtpad{6pt + 1em}\raggedright \hspace{6pt} {\fontsize{8pt}{9.6pt}\selectfont  N/A } \hspace{6pt}\huxbpad{6pt}} &
\multicolumn{1}{l!{\huxvb{0, 0, 0}{0}}}{\cellcolor[RGB]{250, 235, 215}\huxtpad{6pt + 1em}\raggedright \hspace{6pt} {\fontsize{8pt}{9.6pt}\selectfont 2+/-1.5} \hspace{6pt}\huxbpad{6pt}} &
\multicolumn{1}{l!{\huxvb{0, 0, 0}{0}}}{\cellcolor[RGB]{255, 192, 203}\huxtpad{6pt + 1em}\raggedright \hspace{6pt} {\fontsize{8pt}{9.6pt}\selectfont 1.6+/-1.2} \hspace{6pt}\huxbpad{6pt}} &
\multicolumn{1}{l!{\huxvb{0, 0, 0}{0}}}{\cellcolor[RGB]{152, 245, 255}\huxtpad{6pt + 1em}\raggedright \hspace{6pt} {\fontsize{8pt}{9.6pt}\selectfont 1.4+/-1.2} \hspace{6pt}\huxbpad{6pt}} &
\multicolumn{1}{l!{\huxvb{0, 0, 0}{0}}}{\cellcolor[RGB]{255, 192, 203}\huxtpad{6pt + 1em}\raggedright \hspace{6pt} {\fontsize{8pt}{9.6pt}\selectfont 1.6+/-1.3} \hspace{6pt}\huxbpad{6pt}} &
\multicolumn{1}{l!{\huxvb{0, 0, 0}{0}}}{\huxtpad{6pt + 1em}\raggedright \hspace{6pt} {\fontsize{8pt}{9.6pt}\selectfont 1.7+/-1.2} \hspace{6pt}\huxbpad{6pt}} &
\multicolumn{1}{l!{\huxvb{0, 0, 0}{0}}}{\huxtpad{6pt + 1em}\raggedright \hspace{6pt} {\fontsize{8pt}{9.6pt}\selectfont MAE} \hspace{6pt}\huxbpad{6pt}} \tabularnewline[-0.5pt]

\hhline{}
\arrayrulecolor{black}

\multicolumn{1}{!{\huxvb{0, 0, 0}{0}}l!{\huxvb{0, 0, 0}{0}}}{\cellcolor[RGB]{242, 242, 242}\huxtpad{6pt + 1em}\raggedright \hspace{6pt} {\fontsize{8pt}{9.6pt}\selectfont PING-Anx} \hspace{6pt}\huxbpad{6pt}} &
\multicolumn{1}{l!{\huxvb{0, 0, 0}{0}}}{\cellcolor[RGB]{242, 242, 242}\huxtpad{6pt + 1em}\raggedright \hspace{6pt} {\fontsize{8pt}{9.6pt}\selectfont  N/A } \hspace{6pt}\huxbpad{6pt}} &
\multicolumn{1}{l!{\huxvb{0, 0, 0}{0}}}{\cellcolor[RGB]{250, 235, 215}\huxtpad{6pt + 1em}\raggedright \hspace{6pt} {\fontsize{8pt}{9.6pt}\selectfont  1 comp.} \hspace{6pt}\huxbpad{6pt}} &
\multicolumn{1}{l!{\huxvb{0, 0, 0}{0}}}{\cellcolor[RGB]{242, 242, 242}\huxtpad{6pt + 1em}\raggedright \hspace{6pt} {\fontsize{8pt}{9.6pt}\selectfont  N/A } \hspace{6pt}\huxbpad{6pt}} &
\multicolumn{1}{l!{\huxvb{0, 0, 0}{0}}}{\cellcolor[RGB]{255, 192, 203}\huxtpad{6pt + 1em}\raggedright \hspace{6pt} {\fontsize{8pt}{9.6pt}\selectfont  3 comp.} \hspace{6pt}\huxbpad{6pt}} &
\multicolumn{1}{l!{\huxvb{0, 0, 0}{0}}}{\cellcolor[RGB]{152, 245, 255}\huxtpad{6pt + 1em}\raggedright \hspace{6pt} {\fontsize{8pt}{9.6pt}\selectfont  3 comp.} \hspace{6pt}\huxbpad{6pt}} &
\multicolumn{1}{l!{\huxvb{0, 0, 0}{0}}}{\cellcolor[RGB]{242, 242, 242}\huxtpad{6pt + 1em}\raggedright \hspace{6pt} {\fontsize{8pt}{9.6pt}\selectfont  N/A } \hspace{6pt}\huxbpad{6pt}} &
\multicolumn{1}{l!{\huxvb{0, 0, 0}{0}}}{\cellcolor[RGB]{242, 242, 242}\huxtpad{6pt + 1em}\raggedright \hspace{6pt} {\fontsize{8pt}{9.6pt}\selectfont Inferential} \hspace{6pt}\huxbpad{6pt}} \tabularnewline[-0.5pt]

\hhline{}
\arrayrulecolor{black}

\multicolumn{1}{!{\huxvb{0, 0, 0}{0}}l!{\huxvb{0, 0, 0}{0}}}{\huxtpad{6pt + 1em}\raggedright \hspace{6pt} {\fontsize{8pt}{9.6pt}\selectfont PING-Dep} \hspace{6pt}\huxbpad{6pt}} &
\multicolumn{1}{l!{\huxvb{0, 0, 0}{0}}}{\huxtpad{6pt + 1em}\raggedright \hspace{6pt} {\fontsize{8pt}{9.6pt}\selectfont  N/A } \hspace{6pt}\huxbpad{6pt}} &
\multicolumn{1}{l!{\huxvb{0, 0, 0}{0}}}{\cellcolor[RGB]{250, 235, 215}\huxtpad{6pt + 1em}\raggedright \hspace{6pt} {\fontsize{8pt}{9.6pt}\selectfont  0 comp.} \hspace{6pt}\huxbpad{6pt}} &
\multicolumn{1}{l!{\huxvb{0, 0, 0}{0}}}{\huxtpad{6pt + 1em}\raggedright \hspace{6pt} {\fontsize{8pt}{9.6pt}\selectfont  N/A } \hspace{6pt}\huxbpad{6pt}} &
\multicolumn{1}{l!{\huxvb{0, 0, 0}{0}}}{\cellcolor[RGB]{255, 192, 203}\huxtpad{6pt + 1em}\raggedright \hspace{6pt} {\fontsize{8pt}{9.6pt}\selectfont  1 comp. (trend)} \hspace{6pt}\huxbpad{6pt}} &
\multicolumn{1}{l!{\huxvb{0, 0, 0}{0}}}{\cellcolor[RGB]{152, 245, 255}\huxtpad{6pt + 1em}\raggedright \hspace{6pt} {\fontsize{8pt}{9.6pt}\selectfont  1 comp. (trend)} \hspace{6pt}\huxbpad{6pt}} &
\multicolumn{1}{l!{\huxvb{0, 0, 0}{0}}}{\huxtpad{6pt + 1em}\raggedright \hspace{6pt} {\fontsize{8pt}{9.6pt}\selectfont  N/A } \hspace{6pt}\huxbpad{6pt}} &
\multicolumn{1}{l!{\huxvb{0, 0, 0}{0}}}{\huxtpad{6pt + 1em}\raggedright \hspace{6pt} {\fontsize{8pt}{9.6pt}\selectfont Inferential} \hspace{6pt}\huxbpad{6pt}} \tabularnewline[-0.5pt]

\hhline{}
\arrayrulecolor{black}

\multicolumn{1}{!{\huxvb{0, 0, 0}{0}}l!{\huxvb{0, 0, 0}{0}}}{\cellcolor[RGB]{242, 242, 242}\huxtpad{6pt + 1em}\raggedright \hspace{6pt} {\fontsize{8pt}{9.6pt}\selectfont Overall:} \hspace{6pt}\huxbpad{6pt}} &
\multicolumn{1}{l!{\huxvb{0, 0, 0}{0}}}{\cellcolor[RGB]{250, 235, 215}\huxtpad{6pt + 1em}\raggedright \hspace{6pt} {\fontsize{8pt}{9.6pt}\selectfont  6th (worst) } \hspace{6pt}\huxbpad{6pt}} &
\multicolumn{1}{l!{\huxvb{0, 0, 0}{0}}}{\cellcolor[RGB]{242, 242, 242}\huxtpad{6pt + 1em}\raggedright \hspace{6pt} {\fontsize{8pt}{9.6pt}\selectfont  5th } \hspace{6pt}\huxbpad{6pt}} &
\multicolumn{1}{l!{\huxvb{0, 0, 0}{0}}}{\cellcolor[RGB]{242, 242, 242}\huxtpad{6pt + 1em}\raggedright \hspace{6pt} {\fontsize{8pt}{9.6pt}\selectfont  3rd } \hspace{6pt}\huxbpad{6pt}} &
\multicolumn{1}{l!{\huxvb{0, 0, 0}{0}}}{\cellcolor[RGB]{255, 192, 203}\huxtpad{6pt + 1em}\raggedright \hspace{6pt} {\fontsize{8pt}{9.6pt}\selectfont  2nd } \hspace{6pt}\huxbpad{6pt}} &
\multicolumn{1}{l!{\huxvb{0, 0, 0}{0}}}{\cellcolor[RGB]{152, 245, 255}\huxtpad{6pt + 1em}\raggedright \hspace{6pt} {\fontsize{8pt}{9.6pt}\selectfont  1st } \hspace{6pt}\huxbpad{6pt}} &
\multicolumn{1}{l!{\huxvb{0, 0, 0}{0}}}{\cellcolor[RGB]{242, 242, 242}\huxtpad{6pt + 1em}\raggedright \hspace{6pt} {\fontsize{8pt}{9.6pt}\selectfont  4th } \hspace{6pt}\huxbpad{6pt}} &
\multicolumn{1}{l!{\huxvb{0, 0, 0}{0}}}{\cellcolor[RGB]{242, 242, 242}\huxtpad{6pt + 1em}\raggedright \hspace{6pt} {\fontsize{8pt}{9.6pt}\selectfont rank count} \hspace{6pt}\huxbpad{6pt}} \tabularnewline[-0.5pt]

\hhline{}
\arrayrulecolor{black}
\end{tabular}
\end{threeparttable}\par\end{centerbox}

\end{table}

\hypertarget{conclusions}{%
\section{Conclusions}\label{conclusions}}

This paper details SiMLR, an algorithm for computing interpretable
embeddings from high-dimensional, multiple modality datasets. We
demonstrate its performance in simulated data, in two examples on
biomedical data and in one exploratory analysis on public data (in the
supplementary results). SiMLR significantly outperforms RGCCA and its
sparse variant SGCCA in the majority of these results. In a few cases,
the method are statistically equivalent. The methods that we develop are
generalizations of classical methods like PCA and CCA but exist within a
single, consistently implemented framework. The implementation is
efficient for high-dimensional data, builds in network-based
regularization, extends to an arbitrary number of modalities and can be
used for hypothesis testing, clustering or prediction studies. SiMLR was
demonstrated and evaluated in studies relating imaging, genomics,
cognition, demographics and other multi-omic datasets. We provide
strategies for parameter setting, training and testing study design and
the visualization and interpretation of results. The framework is
open-source and relevant to understanding complex, potentially subtle
patterns in healthcare data.

Three supervised learning case studies demonstrate -- with both code and
datasets -- how one may use SiMLR with either covariance related or
regression related similarity measurements and with each source
separation strategy. These examples are designed to show how scientists
may adapt this tool for their own needs. We also recommend default
choices for the similarity (reconstruction energy) and source separation
(ICA) functions with the caveat that the outcomes presented in this
manuscript may be dependent on the internal structure of these example
datasets.

In our simulation study, SiMLR provides more robust signal recovery than
other methods. Because this study is done in a training-testing format,
these results are not due to over-fitting but due to the fact that SiMLR
is able to uncover the ground truth result substantially more
efficiently, as designed. This claim is supported by the fact that
permutation results suggest no predictive signal can be learned (see
online materials in \texttt{codeocean}). We note that both the SVD and
ICA choices for the source separation algorithm achieve overall similar
results. However, the best result is gained by the CCA-like energy
combined with SVD as source separation method.

The simulation study also demonstrates the generative theory under which
SiMLR operates. Consistent with the real biological data examples, SiMLR
outperforms RGCCA/SGCCA in the simulation study. This may suggest that
the simulation strategy is at least to some degree consistent with real
data. Lastly, note that the importance of regularization for solving
inverse problems is well-known at least since
Tikhonov\textsuperscript{18}.

The multi-omic cancer survival prediction study uses SiMLR as a tool for
supervised feature learning. The learned embeddings are linked with
Cox-hazard regression to predict survival in test data. SiMLR's best
performance in this data uses the reconstruction energy with ICA (or
SVD) as source separation method and substantially exceeds SGCCA for
most settings. However, we did note that SGCCA provided solutions with
comparatively good performance for some parameter settings.
SiMLR-Regression-ICA did overall best while SiMLR-CCA-SVD did worst. We
hypothesize that this difference in performance may relate to the fact
that the regression energy refers back to -- in every gradient and
similarity calculation -- the full data matrix. CCA and similar criteria
do not directly use the full data matrices in the energy assessment.
Choice between the ICA and SVD source separation methods did not
substantially impact performance for this problem; the energy function
was overall more important.

These survival prediction outcomes highlight the importance of not only
the selected method but also the data to which it is applied. That is,
the best method for brainAge and in simulation was not the same as for
survival prediction. Lastly, while these models produce average
concordance values up to 0.64, they are not competitive with more
traditional survival models. For example,\textsuperscript{59}
reports concordance results of 0.741, 95\% CI: 0.714--0.768. However,
these cohorts are not identical; more complete data would be required
for a detailed and meaningful comparison.

The PTBP study uses SiMLR to perform supervised dimensionality reduction
across four different datasets, three of which are high-dimensional
neuroimaging and one of which is, to the contrary, a low-dimensional
development-related matrix. The feature vectors are learned in training
data and then applied to brain age prediction in testing data. Relative
to SGCCA, the SiMLR embeddings yield substantially better
cross-validated performance across all settings. We attribute the
advantage to both extra regularization and the direct primal
optimization strategy.

Interestingly, in both simulation and real clinical data, SiMLR extracts
different signal than SGCCA as judged by the systematic performance
trends in these examples. This feature may relate to the method's core
mathematics: high-dimensional embedding vectors are constructed purely
from within modality data but the low-dimensional bases are derived from
cross-modality representations determined by a user-selected source
separation algorithm. If the SVD source separation method is chosen,
then this representation will be orthogonal; if ICA is chosen, they will
be statistically independent where independence is defined by measuring
non-gaussianity\textsuperscript{40} (one of the tenets of fastICA is
``non-gaussianity is independence''). This type of approach will only be
effective in datasets that exhibit some degree of cross-modality
covariation that can be decoded meaningfully into multiple ``true''
source signals. If this is not possible, then SiMLR may obscure rather
than extract hidden signal.

Performance differences could relate to two other implementation
details. SiMLR uses a primal formulation that directly optimizes in the
high-dimensional feature space in which the energy function is defined.
In contrast, SGCCA computes solution updates in a low-dimensional space
(see Algorithm 1 in\textsuperscript{26}) and then performs
soft-thresholding on the resulting vectors after transformation to the
high-dimensional feature space. Secondly, SGCCA uses deflation to
generate multiple components whereas SiMLR operates on full feature
matrices. That is, SiMLR computes full matrix solutions all at once and
uses the underlying source separation method to optimize these vectors
jointly at each iteration of the algorithm. This improves computational
efficiency when extracting several components (i.e.~more than a few) but
also marks a clear difference in the objective functions defined by
SiMLR and SGCCA. These technical factors all contribute to differences
in the outcomes reported here.

There are several limitations to this study and opportunities for future
work. Primarily, we believe this approach and the current findings will
be strengthened by application in related, larger datasets such as those
provided by Adolescent Brain Cognitive Development (ABCD), the UK
Biobank and Human Connectome Project. Furthermore, while we present
methods for matrix standardization (the usual centering and scaling),
this may not be a perfect solution for all cases, in particular when
data deviates strongly from gaussianity. Other alternatives are
available (e.g.~rank transformations), but those are not explored here.
While this work provides several automated or semi-automated strategies
for selecting regularization parameters and the rank (\(k\)) for the
feature vectors, none of these strategies are ``perfect''. This is
unsurprising, given that technical research continues about parameter
setting even in more classical methodology (PCA, CCA). While
cross-validation approaches may also be used, the computational and data
expense for these is relatively high and they also suffer theoretical as
well as practical limitations in terms of
effectiveness\textsuperscript{60}. Despite these issues that are rather
general, we believe the current implementation and interface to SiMLR,
combined with guidance provided here, may yield a practically useful
tool for multiple modality analysis of biomedical imaging and related
data.

A second caveat to this study is that the design is explicitly
multivariate and, as such, we do not interrogate the predictive value of
individual embeddings. Our statistical focus is on the omnibus models.
Other researchers may prefer to study individual embeddings
independently. This is one known limitation within the current
demonstration of SiMLR. Future work may explore this research in
conjunction with extracting not just joint but also individual
structure. This latter advantage is one provided by JIVE. SiMLR could
also be further optimized directly for clustering problems, e.g.~by
implementing a multi-view clustering loss\textsuperscript{61,62}.

Two technical findings from these results are suggestive of directions
for future work. First, SiMLR's performance suggests that a primal
formulation for large joint matrix learning problems is feasible and can
achieve competitive results in real and simulated data. Second, direct
computation of feature matrices (vs.~feature vectors as is done with
deflation schemes) provides computational advantages in our experiments.
However, further analysis of the differences between these technical
approaches within a consistent framework would be needed to draw deeper
conclusions.

Beyond the formulation and implementation of the method, accessibility
of the algorithm is a key contribution. \texttt{ANTsR} is available in
\texttt{GitHub} and via the \texttt{neuroconductor} software
distribution platform. Thus, SiMLR is available for near immediate
access to users who are familiar with the \texttt{R} computing
environment and who wish to test its applicability in their own data. As
always, we recommend interested users contact developers/authors for
guidance or with issues arising in the use of this software.

\hypertarget{acknowledgments}{%
\section{Acknowledgments}\label{acknowledgments}}

This work is supported by the Office of Naval Research N00014-18-1-2440
and K01-ES025432-01. Correspondence and request for materials should be
directed to Dr.~Avants.

\hypertarget{author-contributions}{%
\section{Author contributions}\label{author-contributions}}

B.B.A. and N.J.T. and J.R.S made substantial contributions to the
conception and design of the work and the analysis and interpretation of
data; all authors drafted the work and revised it.

B.B.A. and N.J.T. created the new software used in the work.

\hypertarget{methodology-similarity-driven-multi-view-linear-reconstruction}{%
\section{Methodology: Similarity-driven multi-view linear
reconstruction}\label{methodology-similarity-driven-multi-view-linear-reconstruction}}

\hypertarget{software-platform-antsr}{%
\subsection{\texorpdfstring{Software platform:
\texttt{ANTsR}}{Software platform: ANTsR}}\label{software-platform-antsr}}

The core platform, \texttt{ANTsR}, leverages the powerful \texttt{R}
language to interface and help organize raw neuroimaging, genomics and
other data. \texttt{ANTsR} uses \texttt{Rcpp}\textsuperscript{63} to
wrap Insight ToolKit (ITK, now in version 5\textsuperscript{64}) and
ANTs (currently in version 2.3.3\textsuperscript{65}) \texttt{C++} tools
for the \texttt{R} environment. \texttt{ANTsR} is accessible via both
\texttt{GitHub} and \texttt{neuroconductor}\textsuperscript{66} and is
currently in version v0.5.7.4. Test data and readme files are available
by typing \texttt{?simlr} from within \texttt{ANTsR}. The software used
for this paper is available at \href{https://github.com/ANTsX/ANTsR}{the
\texttt{ANTsR} \texttt{GitHub} repository}.

\hypertarget{technical-background}{%
\subsection{Technical background}\label{technical-background}}

\noindent We discuss, briefly, the primary algorithms upon which SiMLR
is based. We assume data matrices, below, are standardized (columns with
zero mean, unit variance) and \(\| \cdot \|\) denotes the Frobenius
norm.

\hypertarget{multiple-regression}{%
\subsubsection{Multiple regression}\label{multiple-regression}}

\noindent Multiple regression solves a least squares problem that
optimally fits several predictors (the \(n \times p\) matrix \(X\)) to
an outcome (\(y\)). As a quadratic minimization problem, we have:
\[ {\text{arg min}}_{\beta}~\|  y - X
\beta \|^2, \] with optimal least squares solution:
\[ \hat{\beta}=( X^T X)^{-1}
X^T y. \] Above, we may also add a ``ridge'' penalty
\(\lambda \| \beta \| ^2\) on the \(\beta\)s which is useful if
\(p >> n\) i.e.~in the case of complex, multi-view, and multivariate
datasets as we propose to model here. In this document, \(n\) refers to
the number of samples or subjects and \(p\) to predictors.

\hypertarget{principal-component-analysis}{%
\subsubsection{Principal component
analysis}\label{principal-component-analysis}}

\noindent PCA, like multiple regression, may be formulated as the
solution to an energy minimization problem. Select \(k < n\), then find
\(U\) (\(n \times k\)), \(V\) (\(p \times k)\) that minimize
reconstruction error (where we add an \(\ell_1\) constraint as
in\textsuperscript{67--69} to illustrate \emph{sparse} PCA):

\[
{\text{arg min}}_{U,V}~\|  X  - U V^T \|^2   +  \sum_k \lambda_k \| V_k \|_1,
\]

\noindent with additional constraints \(U=XV\) and \(V^T V=I\) where
\(I\) is the identity matrix. The details of these constraints may vary
in regularized variants of the method. Each of the columns of \(X\) is,
here, expressed as a linear combination of the columns of \(U\). For
several modalities, we would compute:
\(\{X_1 = U_1 V_1^T, \cdots ,X_n = U_n V_n^T \}\). In this case, the
``predictors'' are the \(U_i\) and the \(V_i\) is analogous to the
\(\beta\) in the multiple regression case. The \(V_i\) feature vectors
will be sparse if the \(\ell_0\) or \(\ell_1\) penalty is used.

\hypertarget{canonical-correlation-analysis}{%
\subsubsection{Canonical correlation
analysis}\label{canonical-correlation-analysis}}

\noindent CCA may be thought of as a generalization of multiple
regression. Denoting \(Y\) as a \(n \times q\) matrix, CCA seeks to find
solution matrices \(U (k \times p), V (k \times q)\) that maximize
correlation in a low-dimensional space between \(X\) and \(Y\):

\[ {\text{arg max}}_{U,V}~tr( Corr( X U^T, Y V^T ) ), \] where \(Corr\)
is Pearson correlation and \(tr\) is the trace operator. In contrast to
our previous formulation for PCA, CCA evaluates the objective function
(the ``energy'') in a reduced dimensionality space. Any of the methods
above can be made sparse by enforcing the penalties on the feature
weights as described for sparse PCA with the caveat that optimality
constraints must be relaxed. Non-convex optimization methods such as
alternating minimization and/or projected gradient descent must then be
used\textsuperscript{70--72}.

\hypertarget{similarity-driven-multi-view-linear-reconstruction}{%
\subsection{Similarity-driven multi-view linear
reconstruction}\label{similarity-driven-multi-view-linear-reconstruction}}

\noindent SiMLR is a general framework that can be specified in forms
that relate to either sparse PCA (a regression-like objective) or sparse
CCA (a covariance-related objective). The primary concepts are
illustrated in Figure 1. We make two assumptions about datasets to which
we will apply SiMLR.

\begin{itemize}
\item
  \textbf{Assumption 1}: Real latent signal(s) are independent and
  linearly mixed across the biological system on which we are collecting
  several measurements (a standard assumption for blind source
  separation).
\item
  \textbf{Assumption 2}: Sparse, regularized feature vectors can relate
  estimated latent signals in assumption 1 to the original data matrices
  through linear operations.
\end{itemize}

\noindent If data matches these assumptions then methods that can
combine modalities have a better chance of finding the latent signals;
e.g.~joint analysis from (for example) genetics, neuroimaging and
cognition may provide more reliable recovery of the true latent signal
influencing them all. Furthermore, it is likely that spurious signal
will not be shared across all modalities -- or all elements of the
features within a modality -- in a consistent manner. Natural filtering
of noise occurs in joint analysis because (most forms of) noise does not
covary across measurement instances. Adding regularization goes further
in adding robustness: methods regularized with sparseness terms
(\(\ell_0\) or \(\ell_1\)) can down-weight (even to zero) features that
do not improve the objective function. A caveat of these assumptions is
that if no covariation across measurements exists -- or if noise
overwhelms all modalities/measurements -- then these methods may not be
relevant.

\hypertarget{the-simlr-objective-function}{%
\subsubsection{The SiMLR objective
function}\label{the-simlr-objective-function}}

\noindent We first present the high-level framework and will expand upon
details for similarity measurement and regularization below. The core
concepts in SiMLR include the fact that it incorporates flexible
approaches to measuring differences between modalities
(similarity-driven), can take as input several different matrices
(multi-view) and that all operations are linear algebraic in nature
(linear reconstruction). First, we define \(X_i\) as a \(n \times p_i\)
(subjects by features) matrix for a given measurement/view/modality. The
\(i\) ranges from 1 to \(m\) i.e.~the number of modalities (or views).
Then SiMLR optimizes an objective function that seeks to approximate
each modality from its partner matrices through a sparse feature matrix
(\(V_i\)) and low-dimensional representations (\(U_{\ne i}\)):

\[ \begin{aligned} {\text{arg min}}_{V_i} \sum_{i=1}^m S( X_i, f(U_{\ne i}), V_i ) +
\text{Regularization}(V_i),
\end{aligned} \]

\noindent where:

\begin{itemize}
\item
  \(k\) denotes the rank of \(V_i\) and \(U_i\);
\item
  \(V_i\) is a \(p_i \times k\) matrix of feature/solution vectors
  (analogous to \(\beta\)s) for the \(X_i\) modality;
\item
  \(\forall_i ~~U_i = X_i V_i\);
\item
  \(U_{\ne i}\) is a \(n \times ( k ( m - 1 ) )\) low-dimensional
  representation of modalities other than \(X_i\) i.e.~the column-bound
  matrix \(U_{\ne 2}=[U_1,U_3]\) if \(i=2\) and \(m=3\);
\item
  \(f\) is a function (with output dimensionality \(n \times k\)) that
  estimates a low-rank basis set from its argument, is related to
  \textbf{Assumption 1}, and is described in more detail below;
\item
  \(S\) is a function measuring the quality of the approximation of
  \(X_i\) from the other modalities and is related to \textbf{Assumption
  2};
\end{itemize}

\noindent The \(f(U_{\ne i})=\tilde{U}_{\ne i}\) is a key novel
component in the SiMLR framework and is derived by performing blind
source separation over the set of \(j \ne i: \{ X_j V_j \}\) embeddings
(the \(U_{\ne i}\)). We now provide details for each term and other
aspects of the implementation. \newline

\noindent \textbf{Similarity Options.} The default similarity
measurement is one of difference. This is akin to the reconstruction
form for PCA, discussed above. In this case, we have:

\[ S( X_i, \tilde{U}_{\ne i}, V_i ) = \| X_i - \tilde{U}_{\ne i} V_i^T \|^2 . \]

\noindent Here, SiMLR attempts to reconstruct -- in a least-error sense
-- each matrix \(X_i\) directly from the basis representation of the
other \(n-1\) modalities.

We also implement a similarity term inspired by CCA but modified for the
SiMLR objective function. In prior work, we observed that the CCA
criterion -- in the under-constrained form here where we expect
\(p >> n\) -- demonstrates some sensitivity to the sign of
correlations\textsuperscript{73}. As such, we implement an
\emph{absolute canonical covariance (ACC)} similarity measurement
expressed as: \[
\frac{~tr(~|~\tilde{U}_{\ne i}^T X_i V_i ~|~)}{ \|\tilde{U}_{\ne i}\|~ \|X_i
V_i\| }. \] Both reconstruction and ACC have easily computable
analytical derivatives that are amenable to projected gradient descent,
as used in our prior work\textsuperscript{11,12,17}. This similarity
term is most closely related to SABSCOR and SABSCOV in multi-block data
analysis\textsuperscript{74,75}. However, it focuses only on
cross-modality signal.

For the reconstruction energy, SiMLR optimizes these feature vectors to
reconstruct each full matrix from a reduced representation of the other
matrices (the \(\tilde{U}_{\ne i}\)). For ACC, SiMLR optimizes \(V_i\)
to \emph{maximize covariance of \(X_i V_i\) with the low-rank basis}. As
such, the latter similarity term may be more appropriate for recovering
signal that exists more sparsely in the input matrices. \emph{This is
because the operation \(X_i V_i\) is able to completely ignore large
portions of the given matrix \(X_i\)} due to the sparseness terms in our
regularization (described below). The regression energy, on the other
hand, will be more directly informed by the raw high-dimensional matrix
which may have advantages in some cases. Quadratic energies also tend to
have larger capture ranges.

The method's performance also depends on the selection for the basis
representation. We evaluate two options in this initial work:

\begin{itemize}
\tightlist
\item
  \(f_{svd} = svd_u( [U_{\ne i}] )\)
\item
  \(f_{ica} = ica_S( [U_{\ne i}] )\)
\end{itemize}

\noindent The notation \([U_{\ne i}]\) indicates that we bind the
columns together (\texttt{cbind} in \texttt{R}). Below \(alg\) will
represent \(svd_u\) or \(ica_S\). The method \(ica_S\) indicates that we
take the independent components matrix (the \(S\) matrix) from the ICA
algorithm (where ICA produces \(X=AS\)). The method \(svd_u\) indicates
that we take the \(U\) component of the SVD (where SVD produces
\(X=UDV^T\)).

We focus our evaluation on \(ica_S\) and \(svd_u\) functions in this
work as we have found that they produce useful outcomes in example
experiments and they are well-proven methods applied in several domains.
The assumptions underlying ICA and SVD are related. They both fit our
assumption 1 above of linearly mixed independent signals. The difference
is the measure of independence. SVD (or PCA) assumes independence is
measured by variance which leads to orthogonal basis functions. ICA uses
non-gaussianity to measure independence. Both are valid options, from
the theoretical perspective, and we rely on evaluation results to make
recommendations about how to choose between these in practice.

\noindent \textbf{Regularization options.} Regularization occurs on the
\(V_i\) i.e.~our feature matrices. Denote:

\begin{itemize}
\item
  \(v_{ik}\) as the the \(k^\text{th}\) feature vector in \(V_i\);
\item
  \(G_i\) is \(p_i \times p_i\) a sparse regularization matrix with rows
  that sum to one;
\item
  \(\gamma_i\) as a scalar weight which could be used to regularize each
  component differently; effectively, this controls the sparseness and
  varies in zero to one.
\end{itemize}

\noindent Then the regularization terms take the form:

\[ \text{Regularization}(V_i) = \sum_i \sum_k \gamma_{i} \| G_i  v_{ik}
\|^+_{\ell_p}, \]

\noindent where \(\| \cdot \|^+_{\ell_p}\) is the positivity constrained
\(\ell_p\) norm (usually, \(p=0\) or \(p=1\)). This term both enforces
sparseness via \(\ell_p\) while providing data-adaptive degrees of
smoothing via the the graph regularization matrix \(G^i\). For
neuroimaging, this latter feature means that one does not need to
pre-smooth images before running SiMLR. In practice,
\(\| \cdot \|^+_{\ell_p}\) induces unsigned feature vectors. I.e. all
non-zero entries will be either only positive or only negative. \newline

\noindent \textbf{Regularization weights:} The parameterization of the
sparseness for each modality is set by \(\gamma_i\) in the range of zero
to one, where higher values are increasingly sparse (more values of the
feature vector are zero). By default, \(\gamma_i\) is automatically set
to accept the largest 50\(^{th}\) percentile weights but the user may
decide to increase or decrease this value depending on the needs of a
specific study. Alternatively, one may use hyperparameter tuning methods
to automatically determine \(\gamma_i\). For most applications, we
recommend default values.\newline

\noindent \textbf{Regularization matrices:} optional \(G_i\) are
currently set by the user and must be determined in a
data/application/hypothesis-specific manner. In implementation, we
provide helper functions that allow the user to employ \(k\)-nearest
neighbors (KNN) to set the (potentially large) regularization matrices.
We use HNSW\textsuperscript{76} to compute sparse KNN matrix
representations for the \(G_i\). HNSW is among the most efficient
methods currently available and, combined with sparse matrix
representations, make graph regularization on large input matrices
efficient. This aspect of regularization promotes smooth feature vectors
where the nature of smoothness is typically determined by proximity
either spatially or in terms of feature magnitude or feature
correlation.

Although we provide default methods, choice of regularization should
involve some consideration on the part of the user. Because there is no
single theoretically justified answer to these questions, the best
general approach would be to use hyper-parameter optimization.
Alternatively, domain-specific knowledge may be used to guide parameter
setting, in particular sparseness and regularization. Rules of thumb
should be, for regularization, that the estimated \(V_i\) should appear
to reflect biologically plausible feature sets. For sparseness,
biological plausibility should also be considered although we believe
our default parameters provide good general performance. As such,
regularization (i.e.~construction of the \(G_i\)) should perhaps be
given more domain-dependent attention by users. Examples below provide
clarity on how we set these terms in practice. E.g. in neuroimaging, we
may use \(k=5^d\) mask-constrained neighbors for KNN where \(d\) is
image dimensionality. For genomics or psychometrics data, we may set
regularization simply by thresholding correlation (or
\href{https://en.wikipedia.org/wiki/Linkage_disequilibrium}{linkage
disequilibrium}\textsuperscript{77}) matrices.

\hypertarget{simlr-optimization}{%
\subsection{SiMLR: Optimization}\label{simlr-optimization}}

\noindent The overall approach to optimizing the SiMLR objective is that
of projected gradient descent\textsuperscript{78}. In this context, one
derives the optimization algorithm without regularization constraints
and then, at each iteration, projects to the sub-space defined by the
regularization terms. The SiMLR objective function for \(V_i\), at a
given iteration, depends only on the set values for \(X_i\) and
\(\tilde{U}_{\ne i}\). As such, we only need the gradient of the
similarity term with respect to \(V_i\) which greatly simplifies
implementation. We optimize total energy \(E\) via a projected gradient
descent algorithm:

\[ \begin{aligned} \text{loop until convergence:} \\ ~~~~~~~~~~\forall_i
V_i^\text{new} \leftarrow H(~
  G_i~\star~( V_i - \partial S/\partial V_i \epsilon_i~)~) \\
~~~~~~~~~~\forall_i  \tilde{U}_{j \ne i} \leftarrow ~f_{alg}( ~ [~ X_j
V_j^\text{new}~]_{\ne i }~ ) \end{aligned} \]

\noindent where:

\begin{itemize}
\item
  \([~ X_j V_j^\text{new}~]_{\ne i}\) is the collection of
  low-dimensional projections resulting from multiplying the feature
  vectors onto the data matrices where \(\ne i\) indicates that the
  \(i^\text{th}\) projection is held out;
\item
  \(H\) is the thresholding operation which here is applied separately
  to each column of \(V_i\) (see the \emph{iterative hard/soft
  thresholding} literature\textsuperscript{72} and\textsuperscript{78}
  which suggests that \(\ell_0\) penalties provide greater robustness to
  noise);
\item
  \(\epsilon_i\) is a gradient step parameter determined automatically
  by line-search over the total energy \(E\).
\end{itemize}

\noindent Recall that \(f_{alg}\) is a dimensionality reduction step
that reduces \(U_{\ne i}\) to a \(k\)-column matrix. Here, we provide an
example gradient calculation for our default reconstruction error:

\[ \begin{aligned} S = \| X_i - \tilde{U}_{\ne i} V_i^T \|^2,  \\
    \partial S / \partial V_i  = -2  ( X_i^T - V_i \tilde{U}_{\ne i}^T )
    \tilde{U}_{\ne i}, \end{aligned}
\]

\noindent which allows updating the full \(V_i\) at each gradient step.
SiMLR only allows gradient-based updates that improve the total energy;
these are arrived at by line search over the gradient step size and
means that the objective function (driven primarily by the similarity
term) is improved by the new candidate solution; this process is
iterated until the method reaches a fixed point. A fixed point is --
practically speaking -- a convergent solution. I.e. if we further
iterate the algorithm, the solutions do not change beyond some small
numeric fluctuation.

This strategy also allows SiMLR to work directly on the feature matrices
themselves even when \(p >> n\). When large numbers of components are
being computed, this can lead to a distinct computational advantage in
comparison to deflation methods.

\hypertarget{simlr-parameters-and-initialization}{%
\subsection{SiMLR: Parameters and
initialization}\label{simlr-parameters-and-initialization}}

\noindent We summarize default (recommended) parameters and
preprocessing steps for the methodology.

\begin{itemize}
\item
  Matrix pre-processing is performed automatically. Unless the user
  overrides default behavior, we transform each matrix such that:
  \(\forall X_i : X_i \leftarrow \frac{sc(X_i)}{n p_i}\) where \(sc\)
  denotes scaling and centering applied to the matrix columns.
  Normalizing by \(np\) controls the relative scale of the eigenvalues
  of each matrix.
\item
  Number of components (\(k\)) -- The practice for setting these values
  is very similar to practice in PCA or SVD; it may be determined via
  statistical power considerations, cross-validation or set to be
  \(k=n-1\), one less than the number of subjects. This is a problem
  that is currently under active research\textsuperscript{60}.
\item
  Similarity measurement -- \emph{evaluation and comparison of
  similarity choices is ongoing}. Trade-offs are comparable to choosing
  correlation versus Euclidean distance for vectors and better
  performance may be gained in a data-dependent manner. ACC is faster to
  compute on a per-iteration basis but may require more iterations to
  converge. This latter comment is an empirical observation based on our
  studies which, again, may be data dependent.
\item
  Source separation algorithm -- Trade-offs are comparable to choosing
  between SVD and ICA in general. Effectively, ICA should force the
  multiple component solutions toward statistical independence in a
  non-gaussian sense. SVD would be more appropriate for separating
  purely gaussian sources that are mixed linearly.
\end{itemize}

The nature of the feature space is impacted by the constraints on the
\(U_i\) which are determined by the user-selected source separation
algorithm. SVD produces an orthogonal latent space whereas ICA does not.
ICA seeks a latent space that demonstrates statistical independence,
that is, that are maximally non-gaussian\textsuperscript{40,41}. It is
an empirical question about which is ``best'' for a given dataset;
neither is right or wrong in an absolute sense. SVD and ICA are both
used in many practical applications in machine learning and statistics.
Our experiments confirm that both options can produce results that
outperform RGCCA. Overall, the reconstruction error with ICA source
separation appears to give good general performance in our experiments.

SiMLR may be initialized with several different approaches:

\begin{itemize}
\item
  random matrices for all or for each individual modality;
\item
  a joint ICA or SVD across concatenated modalities (recommended and
  default behavior);
\item
  Any other initial low-rank basis set e.g.~derived from RGCCA, etc
  which may be passed to the algorithm by the user;
\end{itemize}

\noindent Due to the fact that SiMLR cannot guarantee convergence to a
global optimum (sparse selection is a NP-hard problem), several
different starting points should be evaluated when using SiMLR in new
problems. This is in concordance with the theory of \texttt{multi-start}
global optimization which we can only approximate in
practice\textsuperscript{79,80}. Other joint reduction methods such as
SGCCA suffer the same limitation. Our recommended default behavior
avoids forcing users to explore multiple starting points but does not
eliminate this fundamental issue that is general to the field of feature
selection in high-dimensional spaces.

\clearpage

\hypertarget{references}{%
\section*{References}\label{references}}
\addcontentsline{toc}{section}{References}

\hypertarget{refs}{}
\leavevmode\hypertarget{ref-Cole2019}{}%
1. Cole, J. H., Marioni, R. E., Harris, S. E. \& Deary, I. J. Brain age
and other bodily `ages': implications for neuropsychiatry. (2019)
doi:\href{https://doi.org/10.1038/s41380-018-0098-1}{10.1038/s41380-018-0098-1}.

\leavevmode\hypertarget{ref-Wray2018}{}%
2. Wray, N. R. \emph{et al.} Genome-wide association analyses identify
44 risk variants and refine the genetic architecture of major
depression. \emph{Nature Genetics} (2018)
doi:\href{https://doi.org/10.1038/s41588-018-0090-3}{10.1038/s41588-018-0090-3}.

\leavevmode\hypertarget{ref-Habeck:2010aa}{}%
3. Habeck, C., Stern, Y. \& Alzheimer's Disease Neuroimaging Initiative.
Multivariate data analysis for neuroimaging data: overview and
application to Alzheimer's disease. \emph{Cell Biochem Biophys}
\textbf{58}, 53--67 (2010).

\leavevmode\hypertarget{ref-Shamy:2011aa}{}%
4. Shamy, J. L. \emph{et al.} Volumetric correlates of spatiotemporal
working and recognition memory impairment in aged rhesus monkeys.
\emph{Cereb Cortex} \textbf{21}, 1559--1573 (2011).

\leavevmode\hypertarget{ref-McKeown:1998aa}{}%
5. McKeown, M. J. \emph{et al.} Analysis of fMRI data by blind
separation into independent spatial components. \emph{Hum Brain Mapp}
\textbf{6}, 160--188 (1998).

\leavevmode\hypertarget{ref-Calhoun:2001aa}{}%
6. Calhoun, V. D., Adali, T., Pearlson, G. D. \& Pekar, J. J. A method
for making group inferences from functional \{MRI\} data using
independent component analysis. \emph{Hum Brain Mapp} \textbf{14},
140--151 (2001).

\leavevmode\hypertarget{ref-Calhoun:2009aa}{}%
7. Calhoun, V. D., Liu, J. \& Adali, T. A review of group \{ICA\} for
f\{MRI\} data and \{ICA\} for joint inference of imaging, genetic, and
\{ERP\} data. \emph{Neuroimage} \textbf{45}, S163--72 (2009).

\leavevmode\hypertarget{ref-Avants2018}{}%
8. Avants, B. \emph{Relating high-dimensional structural networks to
resting functional connectivity with sparse canonical correlation
analysis for neuroimaging}. vol. 136 (2018).

\leavevmode\hypertarget{ref-Pierrefeu2018}{}%
9. Pierrefeu, A. de \emph{et al.} Structured Sparse Principal Components
Analysis With the TV-Elastic Net Penalty. \emph{IEEE transactions on
medical imaging} \textbf{37}, 396--407 (2018).

\leavevmode\hypertarget{ref-Du2016}{}%
10. Du, L. \emph{et al.} Structured sparse canonical correlation
analysis for brain imaging genetics: an improved GraphNet method.
\emph{Bioinformatics (Oxford, England)} \textbf{32}, 1544--1551 (2016).

\leavevmode\hypertarget{ref-Avants2010}{}%
11. Avants, B. \emph{et al.} \emph{Sparse unbiased analysis of
anatomical variance in longitudinal imaging}. vol. 6361 LNCS (2010).

\leavevmode\hypertarget{ref-Avants2014b}{}%
12. Avants, B. \emph{et al.} Sparse canonical correlation analysis
relates network-level atrophy to multivariate cognitive measures in a
neurodegenerative population. \emph{NeuroImage} \textbf{84}, (2014).

\leavevmode\hypertarget{ref-Du2015}{}%
13. Du, L. \emph{et al.} GN-SCCA: Graphnet based sparse canonical
correlation analysis for brain imaging genetics. in \emph{Lecture notes
in computer science (including subseries lecture notes in artificial
intelligence and lecture notes in bioinformatics)} (2015).
doi:\href{https://doi.org/10.1007/978-3-319-23344-4_27}{10.1007/978-3-319-23344-4\_27}.

\leavevmode\hypertarget{ref-Guigui2019}{}%
14. Guigui, N. \emph{et al.} Network regularization in imaging genetics
improves prediction performances and model interpretability on
Alzheimer's disease. in \emph{Proceedings - international symposium on
biomedical imaging} (2019).
doi:\href{https://doi.org/10.1109/ISBI.2019.8759593}{10.1109/ISBI.2019.8759593}.

\leavevmode\hypertarget{ref-Lee1999}{}%
15. Lee, D. D. \& Seung, H. S. Learning the parts of objects by
non-negative matrix factorization. \emph{Nature} (1999)
doi:\href{https://doi.org/10.1038/44565}{10.1038/44565}.

\leavevmode\hypertarget{ref-Chalise2017}{}%
16. Chalise, P. \& Fridley, B. L. Integrative clustering of multi-level
'omic data based on non-negative matrix factorization algorithm.
\emph{PLoS ONE} (2017)
doi:\href{https://doi.org/10.1371/journal.pone.0176278}{10.1371/journal.pone.0176278}.

\leavevmode\hypertarget{ref-Dhillon2014}{}%
17. Dhillon, P. \emph{et al.} Subject-specific functional parcellation
via Prior Based Eigenanatomy. \emph{NeuroImage} \textbf{99}, (2014).

\leavevmode\hypertarget{ref-Tikhonov1943}{}%
18. Tikhonov, A. N. On the stability of inverse problems. \emph{Doklady
Akademii Nauk Sssr} (1943).

\leavevmode\hypertarget{ref-Bell1978}{}%
19. Bell, J. B., Tikhonov, A. N. \& Arsenin, V. Y. Solutions of
Ill-Posed Problems. \emph{Mathematics of Computation} (1978)
doi:\href{https://doi.org/10.2307/2006360}{10.2307/2006360}.

\leavevmode\hypertarget{ref-Smilde2003}{}%
20. Smilde, A. K., Westerhuis, J. A. \& De Jong, S. A framework for
sequential multiblock component methods. \emph{Journal of Chemometrics}
(2003) doi:\href{https://doi.org/10.1002/cem.811}{10.1002/cem.811}.

\leavevmode\hypertarget{ref-Tenenhaus2011}{}%
21. Tenenhaus, A. \& Tenenhaus, M. Regularized Generalized Canonical
Correlation Analysis. \emph{Psychometrika} (2011)
doi:\href{https://doi.org/10.1007/s11336-011-9206-8}{10.1007/s11336-011-9206-8}.

\leavevmode\hypertarget{ref-Tenenhaus2017}{}%
22. Tenenhaus, M., Tenenhaus, A. \& Groenen, P. J. Regularized
Generalized Canonical Correlation Analysis: A Framework for Sequential
Multiblock Component Methods. \emph{Psychometrika} (2017)
doi:\href{https://doi.org/10.1007/s11336-017-9573-x}{10.1007/s11336-017-9573-x}.

\leavevmode\hypertarget{ref-Zhan2018}{}%
23. Zhan, Z., Ma, Z. \& Peng, W. Biomedical Data Analysis Based on
Multi-view Intact Space Learning with Geodesic Similarity Preserving.
\emph{Neural Processing Letters} 1 (2018)
doi:\href{https://doi.org/10.1007/s11063-018-9874-9}{10.1007/s11063-018-9874-9}.

\leavevmode\hypertarget{ref-Baltrusaitis2018}{}%
24. Baltrusaitis, T., Ahuja, C. \& Morency, L. P. Multimodal Machine
Learning: A Survey and Taxonomy. (2018)
doi:\href{https://doi.org/10.1109/TPAMI.2018.2798607}{10.1109/TPAMI.2018.2798607}.

\leavevmode\hypertarget{ref-Kettenring1971}{}%
25. Kettenring, J. R. Canonical analysis of several sets of variables.
\emph{Biometrika} (1971)
doi:\href{https://doi.org/10.1093/biomet/58.3.433}{10.1093/biomet/58.3.433}.

\leavevmode\hypertarget{ref-Tenenhaus2014}{}%
26. Tenenhaus, A. \emph{et al.} Variable selection for generalized
canonical correlation analysis. \emph{Biostatistics} (2014)
doi:\href{https://doi.org/10.1093/biostatistics/kxu001}{10.1093/biostatistics/kxu001}.

\leavevmode\hypertarget{ref-Rohart2017}{}%
27. Rohart, F., Gautier, B., Singh, A. \& Lê Cao, K. A. mixOmics: An R
package for `omics feature selection and multiple data integration.
\emph{PLoS Computational Biology} (2017)
doi:\href{https://doi.org/10.1371/journal.pcbi.1005752}{10.1371/journal.pcbi.1005752}.

\leavevmode\hypertarget{ref-Garali2017}{}%
28. Garali, I. \emph{et al.} A strategy for multimodal data integration:
Application to biomarkers identification in spinocerebellar ataxia.
\emph{Briefings in Bioinformatics} (2017)
doi:\href{https://doi.org/10.1093/bib/bbx060}{10.1093/bib/bbx060}.

\leavevmode\hypertarget{ref-Gloaguen2020}{}%
29. Arnaud Gloaguen , Cathy Philippe, Vincent Frouin, Giulia Gennari,
Ghislaine Dehaene-Lambertz, Laurent Le Brusquet, A. T. Multiway
Generalized Canonical Correlation Analysis. \emph{Biostatistics}
\textbf{In Press}, (2020).

\leavevmode\hypertarget{ref-Hotelling:1935aa}{}%
30. Hotelling, H. Canonical Correlation Analysis (CCA). \emph{J. Educ.
Psychol.} (1935).

\leavevmode\hypertarget{ref-Hotelling:1936aa}{}%
31. Hotelling, H. Relations between two sets of variants.
\emph{Biometrika} 321--377 (1936).

\leavevmode\hypertarget{ref-Lock2013}{}%
32. Lock, E. F., Hoadley, K. A., Marron, J. S. \& Nobel, A. B. Joint and
individual variation explained (JIVE) for integrated analysis of
multiple data types. \emph{Annals of Applied Statistics} (2013)
doi:\href{https://doi.org/10.1214/12-AOAS597}{10.1214/12-AOAS597}.

\leavevmode\hypertarget{ref-Yu2017}{}%
33. Yu, Q., Risk, B. B., Zhang, K. \& Marron, J. S. JIVE integration of
imaging and behavioral data. \emph{NeuroImage} (2017)
doi:\href{https://doi.org/10.1016/j.neuroimage.2017.02.072}{10.1016/j.neuroimage.2017.02.072}.

\leavevmode\hypertarget{ref-Ceulemans2016}{}%
34. Ceulemans, E., Wilderjans, T. F., Kiers, H. A. \& Timmerman, M. E.
MultiLevel simultaneous component analysis: A computational shortcut and
software package. \emph{Behavior Research Methods} (2016)
doi:\href{https://doi.org/10.3758/s13428-015-0626-8}{10.3758/s13428-015-0626-8}.

\leavevmode\hypertarget{ref-Argelaguet2018}{}%
35. Argelaguet, R. \emph{et al.} Multi‐Omics Factor Analysis---a
framework for unsupervised integration of multi‐omics data sets.
\emph{Molecular Systems Biology} (2018)
doi:\href{https://doi.org/10.15252/msb.20178124}{10.15252/msb.20178124}.

\leavevmode\hypertarget{ref-carmichael2019joint}{}%
36. Carmichael, I. \emph{et al.} Joint and individual analysis of breast
cancer histologic images and genomic covariates. \emph{arXiv preprint
arXiv:1912.00434} (2019).

\leavevmode\hypertarget{ref-McMillan2013}{}%
37. McMillan, C. \emph{et al.} White matter imaging helps dissociate tau
from TDP-43 in frontotemporal lobar degeneration. \emph{Journal of
Neurology, Neurosurgery and Psychiatry} \textbf{84}, (2013).

\leavevmode\hypertarget{ref-McMillan2014}{}%
38. McMillan, C. \emph{et al.} Genetic and neuroanatomic associations in
sporadic frontotemporal lobar degeneration. \emph{Neurobiology of Aging}
\textbf{35}, (2014).

\leavevmode\hypertarget{ref-Cook2014}{}%
39. Cook, P. \emph{et al.} Relating brain anatomy and cognitive ability
using a multivariate multimodal framework. \emph{NeuroImage}
\textbf{99}, (2014).

\leavevmode\hypertarget{ref-Hyvarinen1999}{}%
40. Hyvärinen, A. \& Oja, E. Independent component analysis: A tutorial.
\emph{Notes for International Joint Conference on Neural Networks
(IJCNN'99), Washington DC} (1999).

\leavevmode\hypertarget{ref-Hyvarinen2000}{}%
41. Hyvärinen, A. \& Oja, E. Independent component analysis: Algorithms
and applications. \emph{Neural Networks} (2000)
doi:\href{https://doi.org/10.1016/S0893-6080(00)00026-5}{10.1016/S0893-6080(00)00026-5}.

\leavevmode\hypertarget{ref-DeVito2019}{}%
42. De Vito, R., Bellio, R., Trippa, L. \& Parmigiani, G. Multi-study
factor analysis. \emph{Biometrics} (2019)
doi:\href{https://doi.org/10.1111/biom.12974}{10.1111/biom.12974}.

\leavevmode\hypertarget{ref-Haykin2005}{}%
43. Haykin, S. \& Chen, Z. The cocktail party problem. (2005)
doi:\href{https://doi.org/10.1162/0899766054322964}{10.1162/0899766054322964}.

\leavevmode\hypertarget{ref-Goodwin2016}{}%
44. Goodwin, S., McPherson, J. D. \& McCombie, W. R. Coming of age: Ten
years of next-generation sequencing technologies. (2016)
doi:\href{https://doi.org/10.1038/nrg.2016.49}{10.1038/nrg.2016.49}.

\leavevmode\hypertarget{ref-Yong2016}{}%
45. Yong, W. S., Hsu, F. M. \& Chen, P. Y. Profiling genome-wide DNA
methylation. (2016)
doi:\href{https://doi.org/10.1186/s13072-016-0075-3}{10.1186/s13072-016-0075-3}.

\leavevmode\hypertarget{ref-Ozsolak2011}{}%
46. Ozsolak, F. \& Milos, P. M. RNA sequencing: Advances, challenges and
opportunities. (2011)
doi:\href{https://doi.org/10.1038/nrg2934}{10.1038/nrg2934}.

\leavevmode\hypertarget{ref-Andersen1982}{}%
47. Andersen, P. K. \& Gill, R. D. Cox's Regression Model for Counting
Processes: A Large Sample Study. \emph{The Annals of Statistics} (1982)
doi:\href{https://doi.org/10.1214/aos/1176345976}{10.1214/aos/1176345976}.

\leavevmode\hypertarget{ref-Fox2011}{}%
48. Fox, J. \& Weisberg, S. Cox Proportional-Hazards Regression for
Survival Data in R. \emph{Most} (2011).

\leavevmode\hypertarget{ref-Huang2019}{}%
49. Huang, L. \emph{et al.} Development and validation of a prognostic
model to predict the prognosis of patients who underwent chemotherapy
and resection of pancreatic adenocarcinoma: A large international
population-based cohort study. \emph{BMC Medicine} (2019)
doi:\href{https://doi.org/10.1186/s12916-019-1304-y}{10.1186/s12916-019-1304-y}.

\leavevmode\hypertarget{ref-Neums2020}{}%
50. Neums, L., Meier, R., Koestler, D. C. \& Thompson, J. A. Improving
survival prediction using a novel feature selection and feature
reduction framework based on the integration of clinical and molecular
data. in \emph{Pacific symposium on biocomputing} (2020).
doi:\href{https://doi.org/10.1142/9789811215636_0037}{10.1142/9789811215636\_0037}.

\leavevmode\hypertarget{ref-Rappoport2018}{}%
51. Rappoport, N. \& Shamir, R. Multi-omic and multi-view clustering
algorithms: review and cancer benchmark. \emph{Nucleic Acids Research}
(2018)
doi:\href{https://doi.org/10.1093/nar/gky889}{10.1093/nar/gky889}.

\leavevmode\hypertarget{ref-Witten2009}{}%
52. Witten, D. M., Tibshirani, R. \& Hastie, T. A penalized matrix
decomposition, with applications to sparse principal components and
canonical correlation analysis. \emph{Biostatistics} (2009)
doi:\href{https://doi.org/10.1093/biostatistics/kxp008}{10.1093/biostatistics/kxp008}.

\leavevmode\hypertarget{ref-Barnhart2002}{}%
53. Barnhart, H. X., Haber, M. \& Song, J. Overall concordance
correlation coefficient for evaluating agreement among multiple
observers. (2002)
doi:\href{https://doi.org/10.1111/j.0006-341X.2002.01020.x}{10.1111/j.0006-341X.2002.01020.x}.

\leavevmode\hypertarget{ref-Avants2015c}{}%
54. Avants, B. B. \emph{et al.} The pediatric template of brain
perfusion. \emph{Scientific data} (2015)
doi:\href{https://doi.org/10.1038/sdata.2015.3}{10.1038/sdata.2015.3}.

\leavevmode\hypertarget{ref-Kandel2015}{}%
55. Kandel, B., Wang, D., Detre, J., Gee, J. \& Avants, B. Decomposing
cerebral blood flow MRI into functional and structural components: A
non-local approach based on prediction. \emph{NeuroImage} \textbf{105},
(2015).

\leavevmode\hypertarget{ref-Tustison2014a}{}%
56. Tustison, N. \emph{et al.} Logical circularity in voxel-based
analysis: Normalization strategy may induce statistical bias.
\emph{Human Brain Mapping} \textbf{35}, (2014).

\leavevmode\hypertarget{ref-Franke2019}{}%
57. Franke, K. \& Gaser, C. Ten years of brainage as a neuroimaging
biomarker of brain aging: What insights have we gained? (2019)
doi:\href{https://doi.org/10.3389/fneur.2019.00789}{10.3389/fneur.2019.00789}.

\leavevmode\hypertarget{ref-Jernigan2016}{}%
58. Jernigan, T. L. \emph{et al.} The Pediatric Imaging, Neurocognition,
and Genetics (PING) Data Repository. \emph{NeuroImage} (2016)
doi:\href{https://doi.org/10.1016/j.neuroimage.2015.04.057}{10.1016/j.neuroimage.2015.04.057}.

\leavevmode\hypertarget{ref-Manola2011}{}%
59. Manola, J. \emph{et al.} Prognostic model for survival in patients
with metastatic renal cell carcinoma: Results from the international
kidney cancer working group. \emph{Clinical Cancer Research} (2011)
doi:\href{https://doi.org/10.1158/1078-0432.CCR-11-0553}{10.1158/1078-0432.CCR-11-0553}.

\leavevmode\hypertarget{ref-Bro2008}{}%
60. Bro, R., Kjeldahl, K., Smilde, A. K. \& Kiers, H. A.
Cross-validation of component models: A critical look at current
methods. \emph{Analytical and Bioanalytical Chemistry} (2008)
doi:\href{https://doi.org/10.1007/s00216-007-1790-1}{10.1007/s00216-007-1790-1}.

\leavevmode\hypertarget{ref-Bickel2004}{}%
61. Bickel, S. \& Scheffer, T. Multi-view clustering. in
\emph{Proceedings - fourth ieee international conference on data mining,
icdm 2004} (2004).
doi:\href{https://doi.org/10.1109/ICDM.2004.10095}{10.1109/ICDM.2004.10095}.

\leavevmode\hypertarget{ref-Wang2018}{}%
62. Wang, Y., Wu, L., Lin, X. \& Gao, J. Multiview Spectral Clustering
via Structured Low-Rank Matrix Factorization. \emph{IEEE Transactions on
Neural Networks and Learning Systems} (2018)
doi:\href{https://doi.org/10.1109/TNNLS.2017.2777489}{10.1109/TNNLS.2017.2777489}.

\leavevmode\hypertarget{ref-Eddelbuettel2018}{}%
63. Eddelbuettel, D. \& Balamuta, J. J. Extending R with C++: A Brief
Introduction to Rcpp. \emph{American Statistician} (2018)
doi:\href{https://doi.org/10.1080/00031305.2017.1375990}{10.1080/00031305.2017.1375990}.

\leavevmode\hypertarget{ref-Avants2015b}{}%
64. Avants, B., Johnson, H. \& Tustison, N. Neuroinformatics and the the
insight toolkit. \emph{Frontiers in Neuroinformatics} \textbf{9},
(2015).

\leavevmode\hypertarget{ref-Avants2011}{}%
65. Avants, B. \emph{et al.} A reproducible evaluation of ANTs
similarity metric performance in brain image registration.
\emph{NeuroImage} \textbf{54}, (2011).

\leavevmode\hypertarget{ref-Muschelli2019}{}%
66. Muschelli, J. \emph{et al.} Neuroconductor: An R platform for
medical imaging analysis. \emph{Biostatistics} (2019)
doi:\href{https://doi.org/10.1093/biostatistics/kxx068}{10.1093/biostatistics/kxx068}.

\leavevmode\hypertarget{ref-Zou2006}{}%
67. Zou, H., Hastie, T. \& Tibshirani, R. Sparse principal component
analysis. \emph{Journal of Computational and Graphical Statistics}
(2006)
doi:\href{https://doi.org/10.1198/106186006X113430}{10.1198/106186006X113430}.

\leavevmode\hypertarget{ref-Shen2008}{}%
68. Shen, H. \& Huang, J. Z. Sparse principal component analysis via
regularized low rank matrix approximation. \emph{Journal of Multivariate
Analysis} (2008)
doi:\href{https://doi.org/10.1016/j.jmva.2007.06.007}{10.1016/j.jmva.2007.06.007}.

\leavevmode\hypertarget{ref-Jolliffe2003}{}%
69. Jolliffe, I. T., Trendafilov, N. T. \& Uddin, M. A Modified
Principal Component Technique Based on the LASSO. \emph{Journal of
Computational and Graphical Statistics} (2003)
doi:\href{https://doi.org/10.1198/1061860032148}{10.1198/1061860032148}.

\leavevmode\hypertarget{ref-Lin2007}{}%
70. Lin, C. J. Projected gradient methods for nonnegative matrix
factorization. \emph{Neural Computation} (2007)
doi:\href{https://doi.org/10.1162/neco.2007.19.10.2756}{10.1162/neco.2007.19.10.2756}.

\leavevmode\hypertarget{ref-Jain2013}{}%
71. Jain, P., Netrapalli, P. \& Sanghavi, S. Low-rank matrix completion
using alternating minimization. in \emph{Proceedings of the annual acm
symposium on theory of computing} (2013).
doi:\href{https://doi.org/10.1145/2488608.2488693}{10.1145/2488608.2488693}.

\leavevmode\hypertarget{ref-Blumensath2009}{}%
72. Blumensath, T. \& Davies, M. E. Iterative hard thresholding for
compressed sensing. \emph{Applied and Computational Harmonic Analysis}
(2009)
doi:\href{https://doi.org/10.1016/j.acha.2009.04.002}{10.1016/j.acha.2009.04.002}.

\leavevmode\hypertarget{ref-Pustina2018}{}%
73. Pustina, D., Avants, B., Faseyitan, O. K., Medaglia, J. D. \&
Coslett, H. B. Improved accuracy of lesion to symptom mapping with
multivariate sparse canonical correlations. \emph{Neuropsychologia}
\textbf{115}, 154--166 (2018).

\leavevmode\hypertarget{ref-Hanafi2007}{}%
74. Hanafi, M. PLS Path modelling: Computation of latent variables with
the estimation mode B. \emph{Computational Statistics} (2007)
doi:\href{https://doi.org/10.1007/s00180-007-0042-3}{10.1007/s00180-007-0042-3}.

\leavevmode\hypertarget{ref-Tenenhaus2015}{}%
75. Tenenhaus, A., Philippe, C. \& Frouin, V. Computational Statistics
and Data Analysis Kernel Generalized Canonical Correlation Analysis.
\emph{Computational Statistics and Data Analysis} (2015)
doi:\href{https://doi.org/10.1016/j.csda.2015.04.004}{10.1016/j.csda.2015.04.004}.

\leavevmode\hypertarget{ref-Malkov2020}{}%
76. Malkov, Y. A. \& Yashunin, D. A. Efficient and Robust Approximate
Nearest Neighbor Search Using Hierarchical Navigable Small World Graphs.
\emph{IEEE Transactions on Pattern Analysis and Machine Intelligence}
(2020)
doi:\href{https://doi.org/10.1109/TPAMI.2018.2889473}{10.1109/TPAMI.2018.2889473}.

\leavevmode\hypertarget{ref-Hill1968}{}%
77. Hill, W. G. \& Robertson, A. Linkage disequilibrium in finite
populations. \emph{Theoretical and Applied Genetics} (1968)
doi:\href{https://doi.org/10.1007/BF01245622}{10.1007/BF01245622}.

\leavevmode\hypertarget{ref-Bahmani2013}{}%
78. Bahmani, S. \& Raj, B. A unifying analysis of projected gradient
descent for lp- constrained least squares. \emph{Applied and
Computational Harmonic Analysis} (2013)
doi:\href{https://doi.org/10.1016/j.acha.2012.07.004}{10.1016/j.acha.2012.07.004}.

\leavevmode\hypertarget{ref-Marti2013}{}%
79. Martí, R., Resende, M. G. \& Ribeiro, C. C. Multi-start methods for
combinatorial optimization. \emph{European Journal of Operational
Research} (2013)
doi:\href{https://doi.org/10.1016/j.ejor.2012.10.012}{10.1016/j.ejor.2012.10.012}.

\leavevmode\hypertarget{ref-Song2007}{}%
80. Song, G., Avants, B. \& Gee, J. Multi-start method with prior
learning for image registration. in \emph{Proceedings of the ieee
international conference on computer vision} (2007).
doi:\href{https://doi.org/10.1109/ICCV.2007.4409159}{10.1109/ICCV.2007.4409159}.

\end{document}